# Interpretable machine learning for predicting splitting strength of asphalt concrete: insights from SHAP analysis

Jianglei Xing [1], Xiao Tan [1, 2, *], Dongzhao Jin [3*], Pengwei Guo [4], Yuhuan Wang [5], Huiya Niu [6]

[1] *College of Water Conservancy and Hydropower Engineering, Hohai University, Nanjing, Jiangsu, 210024, China*

[2] *State Key Laboratory of Water Disaster Prevention, Nanjing, Jiangsu, 210024, China*

[3] *Department of Civil, Environmental, and Geospatial Engineering, Michigan Technological University, Houghton, Michigan, 49931, United States*

[4] *Faculty of Civil Engineering and Geosciences, Delft University of Technology, Delft, 2628 CN, Netherlands*

[5] *Department of Civil, Environmental, and Architectural Engineering, University of Colorado Boulder, Boulder, Colorado, 80309, United States*

[5] *Shanghai Research Institute of Building Science Co., Ltd., Shanghai, 201108, China*

* Corresponding to: xiaotan@hhu.edu.cn (Xiao Tan) ， dongj@mtu.edu (Dongzhao Jin)

**Abstract**

This paper presents an interpretable machine-learning framework for predicting the splitting strength (ST) of asphalt concrete and supporting data-driven mixture design. A database consisting of 296 samples was established, and 14 input variables related to asphalt properties, aggregate gradation, and fiber characteristics were selected for modeling. Six machine-learning models, namely TabPFN, ANN, SVR, RF, XGBoost, and LightGBM, were developed and compared. Hyperparameter optimization was performed for five models using NSGA-II, while TabPFN was directly applied with its default configuration. The results show that all six models achieved satisfactory predictive capability, whereas TabPFN delivered the best overall performance on the testing set, with the lowest RMSE of 0.28, MAE of 0.21, MAPE of 18.01%, MAD of 0.14, the highest $R^2$ of 0.88, and the highest composite score of 0.91. SHAP analysis further revealed that nine dominant variables accounted for 92.0% of the total average contribution, among which Ag9.5, FT, Ag4.75, AC, and Du were the most influential. In addition, favorable parameter ranges for improving ST were quantified, such as Ag9.5 < 66.8%, Ag4.75 < 45.0%, AC < 5.4 wt.%, AV < 3.6%, and Du > 134.7 cm. Finally, a GUI platform integrating prediction and SHAP-based explanation was developed to improve the accessibility and practical applicability of the proposed framework.

**Abbreviation List**

| Abbreviation | Full name |
|---|---|
| AC | Asphalt content |
| Ag2.36 | 2.36 mm aggregate passing rate |
| Ag4.75 | 4.75 mm aggregate passing rate |
| Ag9.5 | 9.5 mm aggregate passing rate |
| AV | Air voids |
| Du | Ductility |
| FC | Fiber content |
| FL | Fiber length |
| FT | Fiber type |
| Pe | Penetration |
| SP | Softening point |
| ST | Splitting strength |
| TS | Tensile strength |
| VFA | Voids filled with asphalt |
| VMA | Voids in mineral aggregate |

## 1. Introduction

Asphalt concrete has been extensively applied in transportation and hydraulic engineering, including road pavements [1], airport runways [2], parking areas [3], and embankment dams [4]. This wide application is mainly attributed to its favorable waterproofing ability, convenient maintenance, and economic efficiency [5]-[7]. Nevertheless, under practical service conditions, conventional asphalt concrete remains vulnerable to several problems, such as low-temperature brittleness, high-temperature deformation, and gradual deterioration caused by moisture, repeated loading, aging, and temperature variation [8]-[10]. These factors can accelerate cracking and adversely affect the durability and serviceability of pavement structures. Among the commonly used performance indices, splitting strength (ST) is of particular importance because it can directly characterize the tensile resistance and crack susceptibility of asphalt mixtures. Therefore, establishing reliable approaches for ST prediction and identifying the major factors governing its variation are of clear significance for the evaluation and optimization of asphalt concrete.

Conventional experimental evaluation of asphalt concrete performance is generally associated with high cost and low efficiency, since changes in binder content, aggregate properties, or gradation often require repeated and time-consuming laboratory testing [11]-[16]. To reduce this burden, researchers have developed analytical and empirical approaches to estimate mechanical properties [17]. However, although these methods may achieve satisfactory fitting accuracy in specific cases, their applicability is often constrained by simplified theoretical assumptions and limited adaptability to complex variations in mixture design. Mechanistic-based prediction frameworks have also been employed to assess long-term pavement performance, especially in relation to fatigue and rutting behavior [18]. Nevertheless, the dependence of such approaches on fixed material parameters and predefined structural assumptions reduces their generalization capability when dealing with heterogeneous asphalt mixtures [19].

Machine learning (ML) has recently become an important analytical approach in concrete-materials research [20], [21], particularly for identifying hidden patterns and forecasting material properties from complex datasets. Because ML methods are well suited to large, heterogeneous data and can represent nonlinear correlations among multiple material parameters, they have been increasingly adopted for investigating the mechanical performance of engineering materials [22]. In asphalt concrete, models such as artificial neural networks (ANN), random forests (RF), k-nearest neighbors (KNN), and light gradient boosting machines (LightGBM) [21]-[24], have already been introduced for the prediction of major mechanical properties, and the reported results indicate satisfactory accuracy. For example, RF achieved an $R^2$ of 0.83 in predicting Marshall stability [23], while KNN and LightGBM yielded even higher predictive performance, reaching an $R^2$ of 0.90 [24]. These results indicate that ML-based approaches can provide more flexible and accurate predictions than traditional empirical models, particularly when dealing with diverse mixture compositions and complex variable interactions.

Despite the progress achieved so far, the broader use of machine learning in asphalt concrete research is still hindered by several challenges: the machine-learning models adopted in previous studies are mostly conventional, whereas newer and potentially more powerful models have rarely been explored in asphalt concrete research; (2) existing studies have mainly emphasized prediction accuracy, whereas integrated frameworks that combine model comparison, comprehensive evaluation, and interpretability analysis remain limited, thereby restricting a transparent understanding of how input variables influence splitting strength [25]; and (3) publicly accessible platforms for asphalt concrete splitting strength prediction are still scarce, limiting the practical usability of these methods for engineers and researchers.

To address these issues, an interpretable machine-learning framework was established in this study for predicting the ST of asphalt concrete. A literature-derived dataset incorporating asphalt-related properties, aggregate gradation parameters, and mixture-design variables was used to benchmark six ML models, after which the optimal model was selected through a comprehensive performance assessment. To provide insight into the prediction process, SHapley Additive exPlanations (SHAP) was employed to determine the most influential variables and to characterize how they contributed to ST variation. In addition, a graphical user interface (GUI) was built to combine prediction with interpretation, thereby facilitating practical use of the proposed method.

The principal novelties of this study can be summarized in three aspects: (1) As a newer and more powerful machine-learning paradigm, Tabular Prior-data Fitted Network (TabPFN) was introduced for asphalt concrete ST prediction, and its ability to deliver excellent performance without dataset-specific hyperparameter tuning was verified, highlighting its efficiency and applicability for small-to-medium-sized tabular datasets; (2) an interpretable machine-learning framework was established for asphalt concrete ST prediction by combining data preprocessing, multi-model comparison, composite-indicator-based evaluation, and SHAP analysis, which not only improved predictive reliability but also revealed the relative importance of input variables and their influence patterns on ST; and (3) an intuitive graphical user interface (GUI) platform was developed to integrate prediction and interpretation into an interactive tool for asphalt concrete evaluation and design.

## 2. Methodology

### 2.1. Study framework

**Fig. 1** illustrates the workflow of the proposed interpretable machine-learning framework, which consists of four main steps. (1) Dataset development and preprocessing: a dataset with 296 samples was established using 14 input variables and ST as the output. Mean imputation, one-hot encoding for FT, and Z-score standardization were applied, followed by an 80:20 train-test split. (2) Model development and comparison: six machine-learning models were developed for ST prediction. NSGA-II was used to optimize five conventional models, whereas TabPFN was directly applied with its default settings. Their predictive performance was then evaluated using a comprehensive metric, and TabPFN was identified as the best-performing model. (3) SHAP-based interpretation and knowledge extraction: SHAP analysis was used for local and global interpretation, including feature importance ranking and dependence analysis, to identify favorable parameter ranges for ST improvement. (4) GUI platform: the best-performing model and SHAP interpretation results were integrated into a GUI platform for interactive ST prediction and explanation.

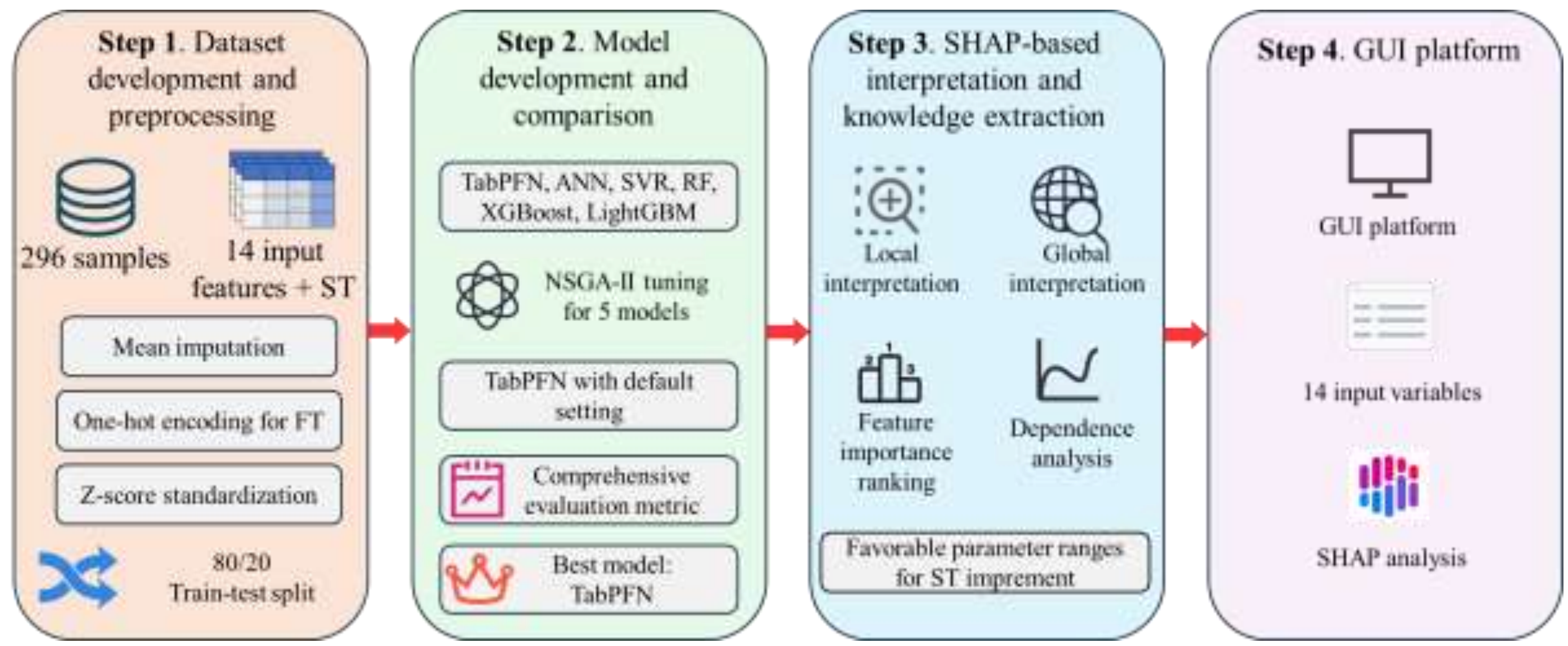

Fig. 1. Workflow of the proposed interpretable machine-learning framework for asphalt concrete splitting strength prediction, SHAP-based interpretation, and GUI application.

## 2.2. Database development

### *2.2.1. Database construction and description*

A database containing 296 asphalt concrete samples was established from relevant studies published between 2008 and 2025 for ST prediction [26]-[56]. Taking into account the material properties of asphalt, aggregate gradation, and fibers, fourteen input variables were selected and classified into three groups to investigate their effects on splitting strength [57]: asphalt-related features, including asphalt content (AC), penetration (Pe), softening point (SP) and ductility (Du); aggregate-related features, including the passing percentages of 2.36 mm, 4.75 mm, and 9.5 mm aggregates (Ag2.36, Ag4.75, and Ag9.5), air voids (AV), voids in mineral aggregate (VMA), and voids filled with asphalt (VFA); and fiber-related features, namely fiber content (FC), fiber type (FT), tensile strength (TS) and fiber length (FL). When the samples are classified according to fiber type, the database includes basalt fiber, glass fiber, polyester fiber, steel fiber, and no fiber, and their distribution proportions are shown in **Table 1**.

Table 1. Overview of fiber types and their amounts

| Fiber types | Sample size |
|---|---|
| Basalt fiber | 17 |
| Glass fiber | 14 |
| Polyester fiber | 20 |
| Steel fiber | 4 |
| No fiber | 241 |

### *2.2.2. Data analysis*

The missing values in the database were filled using mean imputation. Following imputation, the statistical profiles of all variables are summarized in **Table 2**, covering the minimum, maximum, quartile values, mean, and standard deviation for each variable.

Table 2. Distribution of the fourteen input variables and the output variable

| Variable | Unit | Min | Q1 | Q2 | Q3 | Max | Mean | STD |
|---|---|---|---|---|---|---|---|---|
| Pe | 0.1mm | 47 | 63 | 71.2 | 85.9 | 93 | 71.65 | 14.05 |
| Du | cm | 98 | 100 | 101 | 150 | 200 | 125.65 | 31.21 |
| SP | °C | 44.1 | 47.2 | 50 | 57 | 73 | 53 | 7.87 |
| AC | % by mass | 3 | 4.6 | 4.9 | 6.5 | 8 | 5.35 | 1.17 |
| Ag2.36 | % | 13.9 | 26.58 | 32.92 | 40.15 | 56 | 33.54 | 10.9 |
| Ag4.75 | % | 23.9 | 37.9 | 50.77 | 58.89 | 71 | 47.66 | 13.53 |
| Ag9.5 | % | 53 | 62.76 | 76.16 | 81.2 | 86 | 72.88 | 10.3 |
| AV | % | 2.54 | 4.01 | 4.34 | 4.95 | 8 | 4.41 | 0.98 |
| VMA | % | 12.1 | 14.94 | 15.36 | 16.2 | 65.6 | 17.08 | 8.65 |
| VFA | % | 17.11 | 69.03 | 72.95 | 82.59 | 83.41 | 72.91 | 11.81 |
| FC | % | 0 | 0 | 0 | 0 | 3 | 0.09 | 0.36 |
| FT | / | / | / | / | / | / | / | / |
| TS | MPa | 0 | 0 | 0 | 0 | 3250 | 237.17 | 735.13 |
| FL | mm | 0 | 0 | 0 | 0 | 12 | 1.23 | 2.94 |
| ST | MPa | 0.13 | 0.71 | 1.1 | 1.48 | 5.15 | 1.33 | 0.91 |

The probability density characteristics of the fourteen input variables and the output variable are presented in **Fig. A1**. For each feature (such as Pe, Du, and SP), a dual-axis subplot is used, with probability density shown on the left y-axis and frequency shown on the right y-axis. Pearson correlation analysis was further carried out to evaluate possible multicollinearity, and the results are provided in **Fig. 2**. [58]. Results show that, except for a few relatively strong correlations among asphalt, aggregate, and fiber internal features, the correlations among the remaining features all satisfy $|R| < 0.7$ [59]. This indicates weak linear relationships and limited redundancy among the variables, confirming that the selected features were appropriate for model training. Because no severe multicollinearity was observed among the selected variables, all mixture design variables were kept in the input set [60].

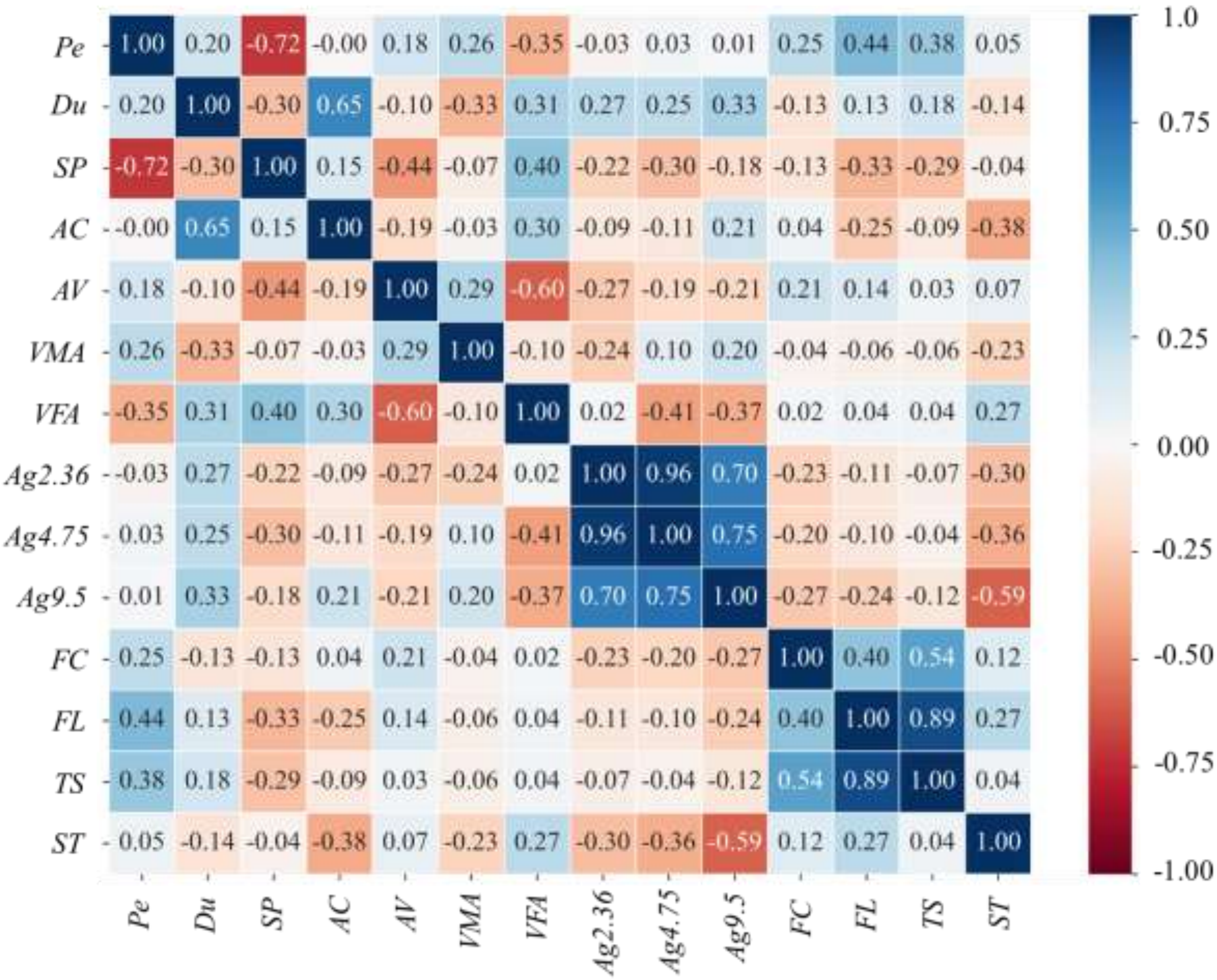

Fig. 2. Pearson correlation heatmap of the dataset.

### *2.2.3. Data preprocessing*

For the categorical variable FT, one-hot encoding was applied to transform it into a representation that could be directly used by the machine-learning models [61]. For the remaining numerical features, standardization was further performed before model training. Data standardization is a necessary procedure in machine learning, as it maps variables with different measurement scales into a common numerical range. This process can improve training efficiency and speed up convergence by reducing the influence of variables with larger magnitudes [62]. In this study, the input variables were standardized using Z-score scaling before machine-learning modeling [63]. The formula was shown as follows:

$$z = \frac{x - m}{\sigma} \tag{1}$$

where $x$ denotes an input variable, $m$ and $\sigma$ represent its mean and standard deviation, respectively.

The standardized dataset was then randomly reordered to minimize possible sequence-related effects and to enhance the representativeness of the training samples [60]. After that, 80% of the samples were assigned to the training set and the remaining 20% to the testing set [64].

**2.3. Machine learning models**

Six machine learning models were employed in this study: Tabular Prior-data Fitted Network (TabPFN) [65], Support Vector Regression (SVR) [66], Random Forest (RF) [67], Extreme Gradient Boosting Trees (XGBoost) [68], Light Gradient Boosting Machine (LightGBM) [69] and Artificial Neural Network (ANN) [70]. The classification of these models is presented in **Table 3**.

Table 3. Overview of the machine learning models

| No. | Model | Category | Notes |
|---|---|---|---|
| 1 | TabPFN | Foundation model | Transformer-based prediction |
| 2 | ANN | Classical | Nonlinear regression |
| 3 | SVR | Classical | Kernel-based regression |
| 4 | RF | Ensemble – Bagging | Bagging of decision trees |
| 5 | XGBoost | Ensemble – Boosting | Boosting model with regularization |
| 6 | LightGBM | Ensemble – Boosting | Efficient histogram-based gradient boosting |

Among the six models, TabPFN deserves particular attention because it differs fundamentally from conventional tabular learning models. As illustrated in **Fig. 3(a)**, TabPFN is a pretrained tabular foundation model that is trained on a large collection of synthetically generated tabular tasks sampled from a broad prior over data-generating processes [71]. Through this pretraining stage, the model learns a general prior for tabular prediction, rather than being trained entirely from scratch for each downstream dataset. This enables TabPFN to capture transferable predictive patterns from many synthetic tabular tasks before being applied to a specific real-world problem.

As shown in **Fig. 3(b)**, in the present study, the fourteen input descriptors of asphalt concrete, including asphalt-related, aggregate-related, and fiber-related variables, were organized into tabular inputs and then fed into the pretrained TabPFN model to predict the target variable ST. Its underlying architecture is Transformer-based, in which tabular inputs are encoded and processed through stacked Transformer blocks before being passed to the prediction head [65]. This design makes TabPFN particularly suitable for the present ST database, which contains a moderate number of samples and a compact set of descriptors, because it reduces the need for repeated dataset-specific retraining and extensive hyperparameter optimization while still maintaining strong predictive performance. In practice, the open-source TabPFN implementation provided by PriorLabs was adopted in this study, and its regressor interface was used to conduct the prediction experiments [72].

**(a) Pre-training on synthetic tabular tasks**

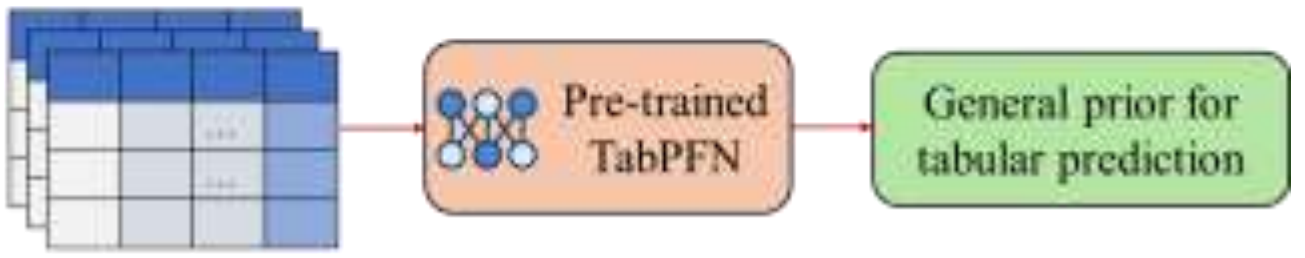

Learned from many synthetic tabular tasks

**(b) Application to asphalt concrete dataset**

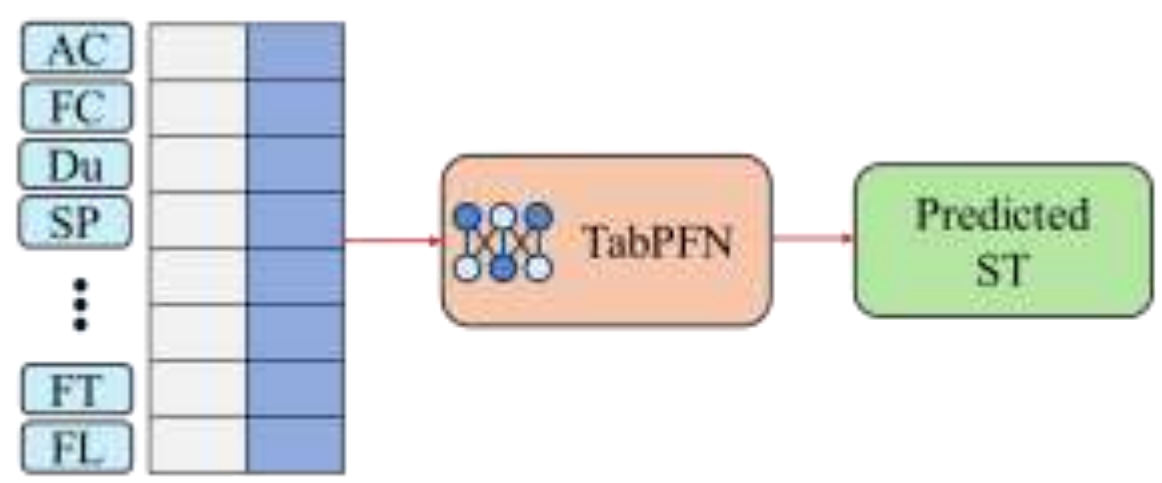

Fig. 3. Schematic diagram of TabPFN pre-training and its application to asphalt concrete ST prediction.

### 2.4. Evaluation metrics

Model performance was evaluated using five indicators: root mean square error (RMSE), mean absolute percentage error (MAPE), mean absolute error (MAE), median absolute deviation (MAD) of the residuals, and the coefficient of determination ($R^2$). Their mathematical definitions are given in Eqs. (2)–(6). Among them, MAE denotes the mean prediction error in the original unit of the target variable, which makes it readily interpretable. RMSE is calculated from the squared differences between predicted and observed values and is more sensitive to large errors. MAPE represents the prediction error in percentage form, which allows comparisons across different scales. MAD measures the median absolute residual and is therefore more robust to extreme errors than mean-based indicators. $R^2$ reflects the proportion of variance in the target variable explained by the model, with values closer to 1 indicating better fitting performance. In general, lower RMSE, MAPE, MAE, and MAD values indicate higher predictive accuracy, whereas a higher $R^2$ suggests superior model performance.

$$RMSE = \sqrt{\frac{1}{n}\sum_{i=1}^{n}(y_{i,pre} - y_{i,test})^2} \tag{2}$$

$$MAPE = \frac{1}{n}\sum_{i=1}^{n}\left|\frac{y_{i,pre} - y_{i,test}}{y_{i,test}}\right| \times 100\% \tag{3}$$

$$MAE = \frac{1}{n}\sum_{i=1}^{n}\left|y_{i,pre} - y_{i,test}\right| \tag{4}$$

$$\begin{cases} e_i = y_i - \hat{y}_i \\ MAD = median(|e_i - median(e)|) \end{cases} \tag{5}$$

$$R^2 = 1 - \frac{\sum_{i=1}^{n}(y_{i,test} - y_{i,pre})^2}{\sum_{i=1}^{n}(y_{i,test} - \bar{y})^2} \tag{6}$$

Note: For each target variable, *n* represents the number of samples in the testing (or training) set,

$y_i$ and $\hat{y}_i$ are the actual and predicted values of the $i$-th sample, respectively; $\bar{y}$ is the mean of the ground-truth values; and $e_i$ represents the residual for the $i$-th sample.

To enable an integrated comparison across metrics with different scales and optimization directions, the five evaluation metrics were further transformed into dimensionless bounded scores within the interval (0,1), where a larger value consistently indicates better performance. Since RMSE, MAPE, MAE, and MAD are error-type metrics for which smaller values are preferred, whereas a larger $R^2$ is preferred, the first step was to convert all metrics into a unified "higher-is-better" utility:

$$u_{k,m} = \begin{cases} -x_{k,m}, & k \in \{RMSE, MAPE, MAE, MAD\} \\ x_{k,m}, & k \in R^2 \end{cases} \tag{7}$$

where $x_{k,m}$ denotes the raw value of metric $k$ for model $m$, and $u_{k,m}$ is the corresponding utility value after directional unification.

Then, for each metric $k$, z-score standardization was performed across all candidate models:

$$z_{k,m} = \frac{u_{k,m} - \mu_k}{\sigma_k} \tag{8}$$

where $\mu_k$ and $\sigma_k$ are the mean and standard deviation of the utility values of metric $k$ across all models, respectively.

Finally, to obtain a stable bounded score and reduce the dominance of extreme values, the standardized values were mapped into the interval (0,1) using a sigmoid function:

$$s_{k,m} = \frac{1}{1 + exp^{(-z_{k,m}/\alpha)}} \tag{9}$$

where $\alpha$ is a scaling parameter controlling the steepness of the transformation. In this study, $\alpha = 0.5$.

Based on the five normalized metric scores, the composite score of model $m$ was calculated as the arithmetic mean of the normalized scores over the five metrics:

$$Composite\ Score_m = \frac{1}{5}\sum_{k\in K} s_{k,m} \quad (10)$$

where

$$K = \{RMSE, MAPE, MAE, MAD, R^2\} \quad (11)$$

A larger composite score indicates better overall predictive performance after jointly considering prediction accuracy, goodness of fit, and residual stability. In this study, the normalized metric scores were used to construct the radar chart in **Fig. 8(a)**, while the composite score was used to rank the overall performance of different models in **Fig. 8(b)**.

**2.5. Hyperparameter tuning through objective optimization**

Unlike conventional machine learning models, TabPFN was not subjected to hyperparameter tuning in this study. This is because TabPFN is designed as a pretrained tabular foundation model, whose predictive capability mainly stems from large-scale prior pretraining rather than dataset-specific parameter adjustment. According to the characteristics of the model itself and the recommendation of its original authors, TabPFN is intended to be used largely in its default configuration, thereby avoiding the expensive and often unnecessary hyperparameter optimization process required by many traditional machine learning algorithms. This property is also one of the practical advantages of TabPFN, especially for small-to-medium-sized tabular datasets, as it allows strong predictive performance to be achieved with minimal manual intervention [65].

For the remaining machine learning models, hyperparameter optimization plays an important role in improving predictive performance. Common approaches include Grid Search [73], Random Search [74] and Genetic Algorithms-based method [75]. Grid Search and Random Search were not adopted in this study because they become computationally inefficient when exploring high-dimensional hyperparameter spaces and are less effective in capturing complex interactions among hyperparameters [76]. Therefore, the Non-dominated Sorting Genetic Algorithm II (NSGA-II), a Genetic Algorithms-based method, was employed. NSGA-II provides an efficient global search strategy by maintaining population diversity and balancing exploration and exploitation during the optimization process [77]. It searches the hyperparameter space through an evolutionary process in which each generation contains multiple candidate solutions, and each individual represents a specific hyperparameter setting. In the present work, the tuning process aimed to minimize the RMSE of the target variable ST. The corresponding tuning procedure is illustrated in **Fig. 4**.

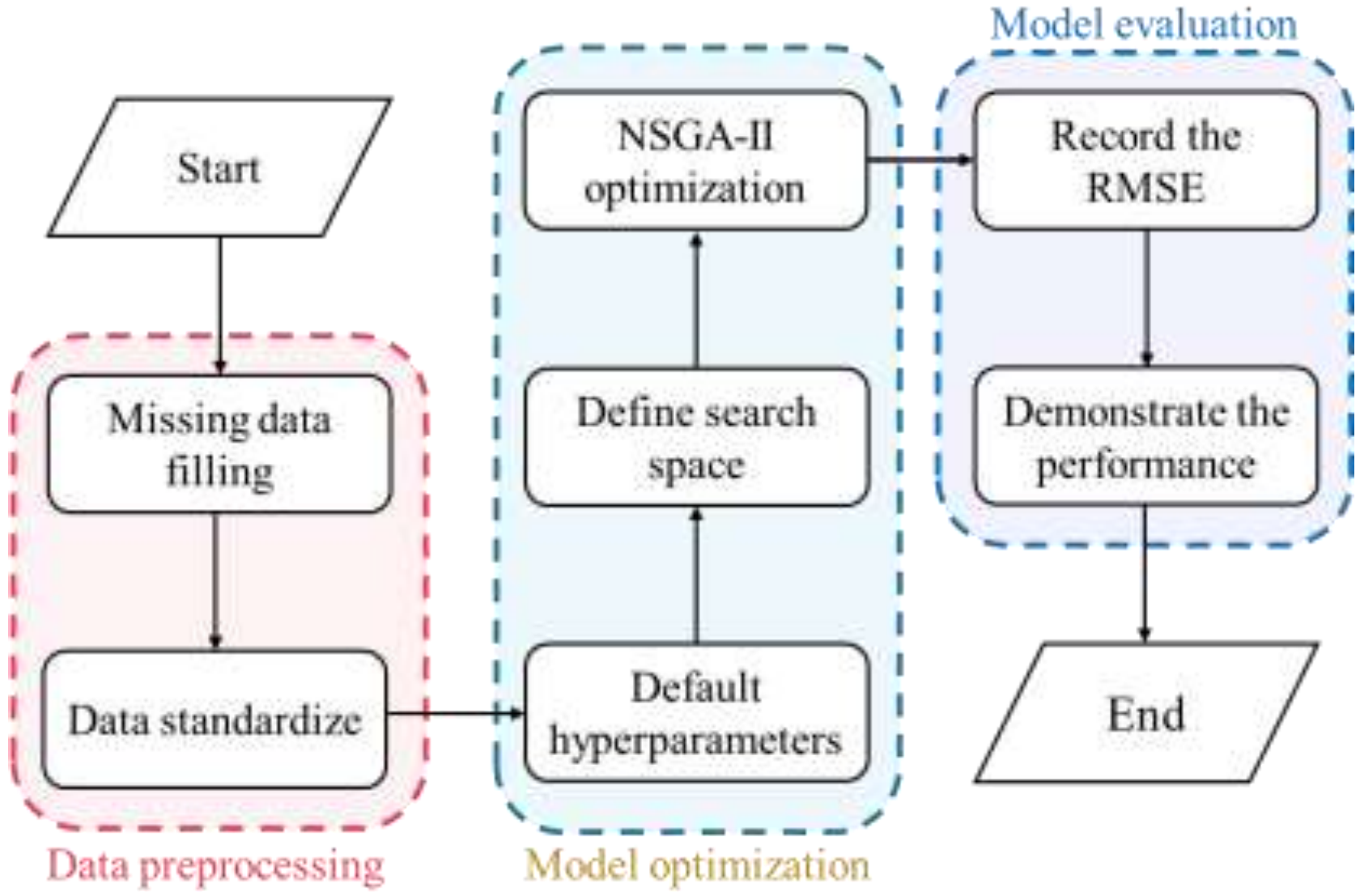

Fig. 4. Flowchart of the NSGA-II-based hyperparameter tuning process.

**2.6. SHAP-based model explanation**

ML models have shown strong predictive capability in materials engineering, yet their practical application is often constrained by limited interpretability because many of them function as “black-box” systems. To address this issue, SHAP [78] can be employed as a post hoc interpretation tool to quantify the contribution of each input variable and thus provide transparent explanations for model outputs. As illustrated in **Fig. 5**, the SHAP-based explanation framework can be understood in four parts. In **Fig. 5(a)**, the input variables are first fed into the trained ML model, which operates as a black-box predictor and generates the target prediction. **Fig. 5(b)** then shows that this prediction can be decomposed into a baseline value together with the contribution of individual input features, expressed as SHAP values. **Fig. 5(c)** further illustrates the additive principle of SHAP, in which the final prediction is obtained by adjusting the base value through the positive or negative effects of different variables. In this process, positive SHAP values increase the prediction, whereas negative SHAP values decrease it. Finally, **Fig. 5(d)** presents the interpretation results at both local and global levels. Local analysis explains how a single prediction is formed, while global analysis summarizes the overall importance and influence patterns of features across the entire dataset. In this way, SHAP converts the original black-box prediction into an interpretable explanation framework, thereby improving the transparency of the model decision-making process.

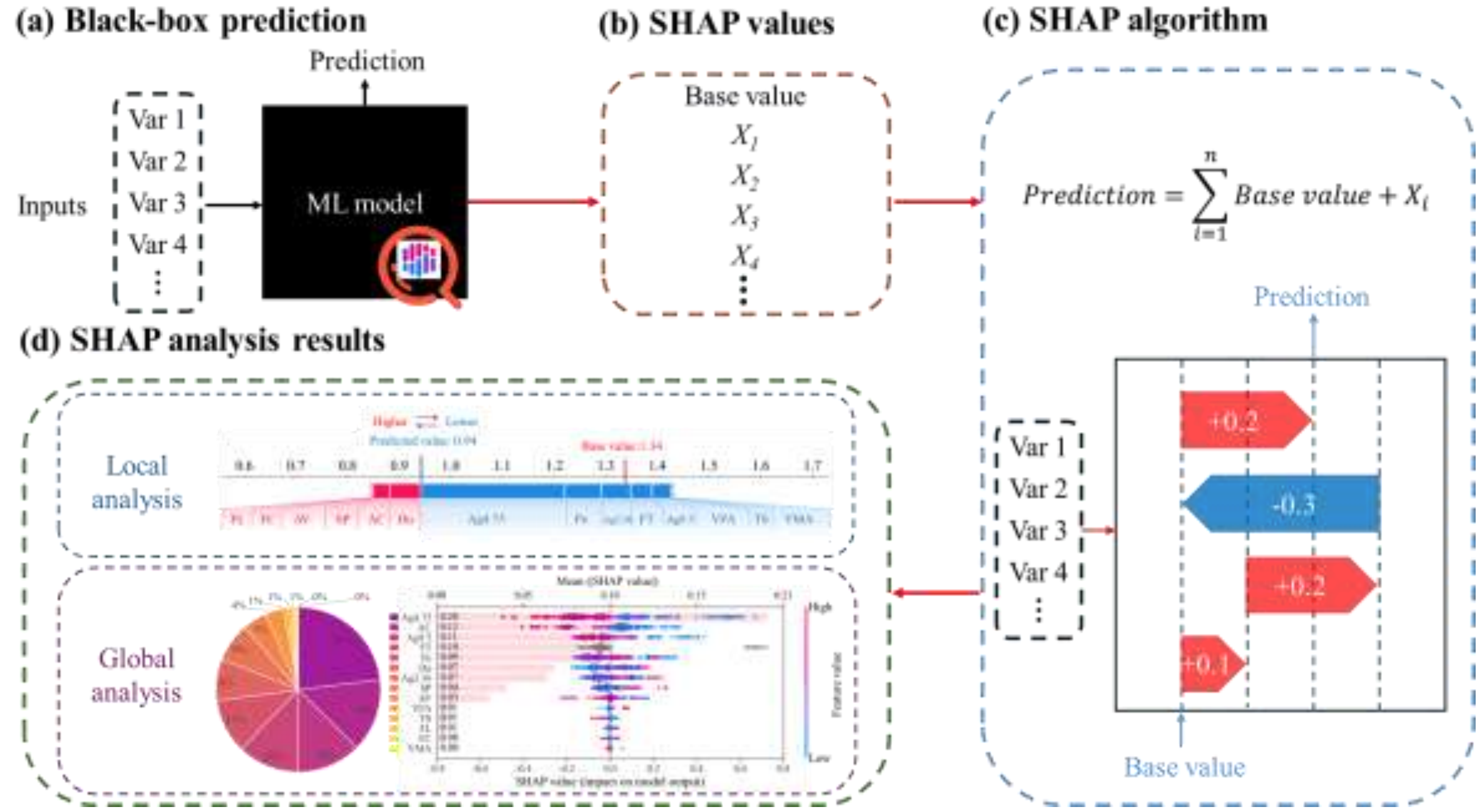

Fig. 5. Flowchart of the SHAP-based interpretation process for machine learning model predictions.

## 3. Results and discussion

### 3.1. Hyperparameter optimization results

Optimal hyperparameter combinations were identified using the pymoo-based NSGA-II algorithm, which was executed with a population size of 16 over 100 generations. For each candidate hyperparameter set, the objective value was defined as the mean RMSE from 5-fold cross-validation on the training data, and the optimization process aimed to minimize this value.

Based on the above optimization settings, the termination criterion was defined as 100 generations. As shown in **Fig. 6**, all five models converged well before the preset maximum generation, indicating that this setting provided sufficient search depth while avoiding unnecessary computational expense. Specifically, RF, XGBoost, and LightGBM reached stable validation RMSE values within the early generations, whereas SVR showed only slight improvement after its initial convergence. By contrast, ANN exhibited a relatively slower optimization process, with a pronounced reduction in RMSE during the early generations and a gradual plateau after approximately 30–40 generations, followed by only marginal improvement thereafter. Overall, **Fig. 6** illustrates the evolution of the best validation RMSE during the NSGA-II optimization process, confirming that the selected generation limit was adequate and that extending the search further would be unlikely to produce substantial performance gains. The resulting hyperparameter combinations for all tuned models are provided in **Appendix B**, **Table B1**.

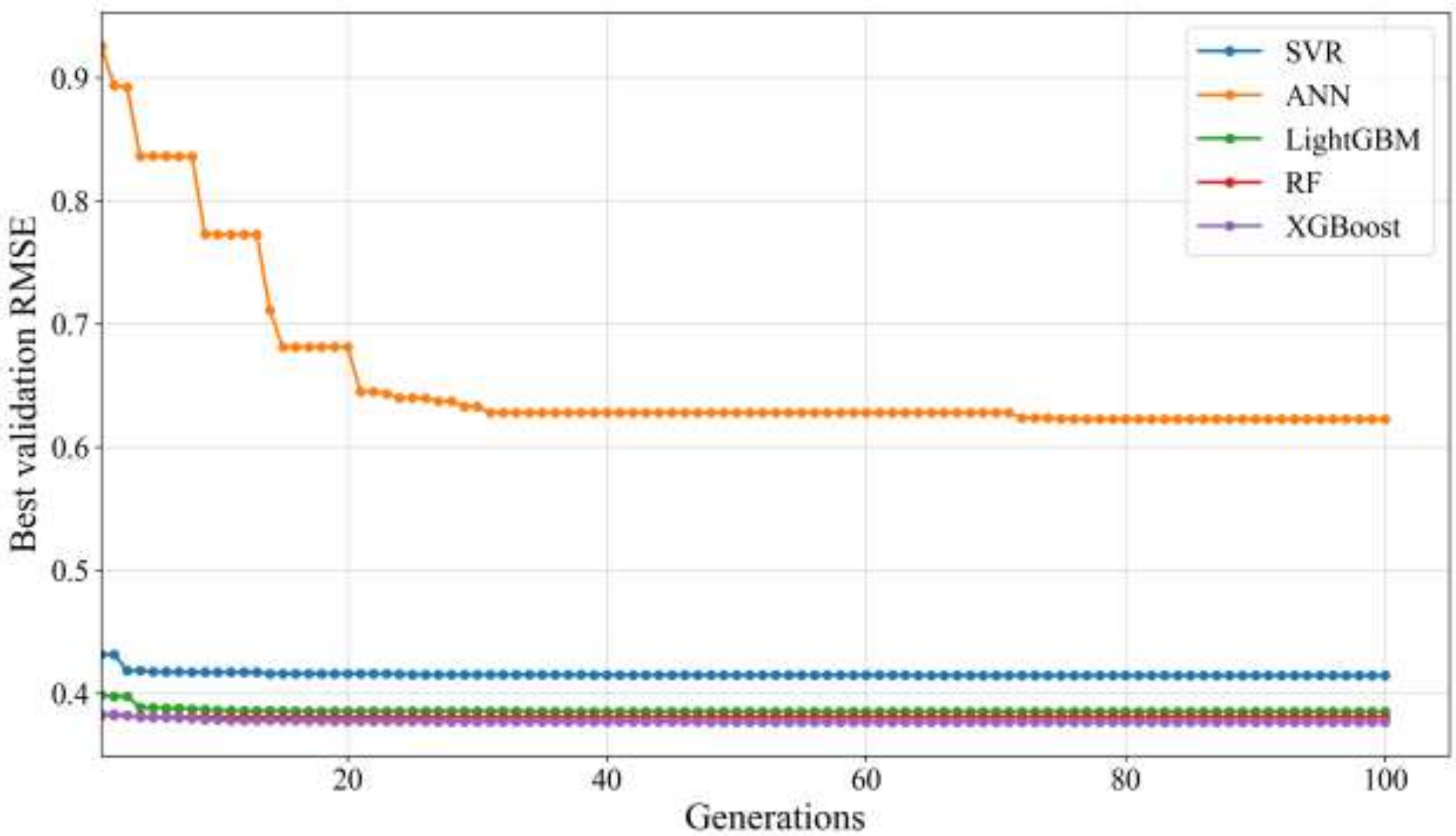

Fig. 6. RMSE variation of the models during NSGA-II iterations.

### 3.2. Prediction performance

**Fig. 7** compares the fitting and generalization behavior of the six machine-learning models for ST prediction using the training and testing datasets. The scatter points for the two datasets are plotted against the 1:1 reference line, which represents perfect agreement between model outputs and measured values. Data points located closer to this reference line indicate stronger predictive consistency and smaller deviations. It is also noted that MAE is lower than RMSE for all models, which is consistent with the general expectation for satisfactory machine-learning predictions [79]. Overall, all six models were capable of predicting ST with satisfactory accuracy. Among them, TabPFN achieved the best overall performance on the testing set, with the lowest RMSE (0.28) and the highest $R^2$ value of 0.88. The detailed performance metrics of all models on the testing set are provided in **Appendix B**, **Table B2**.

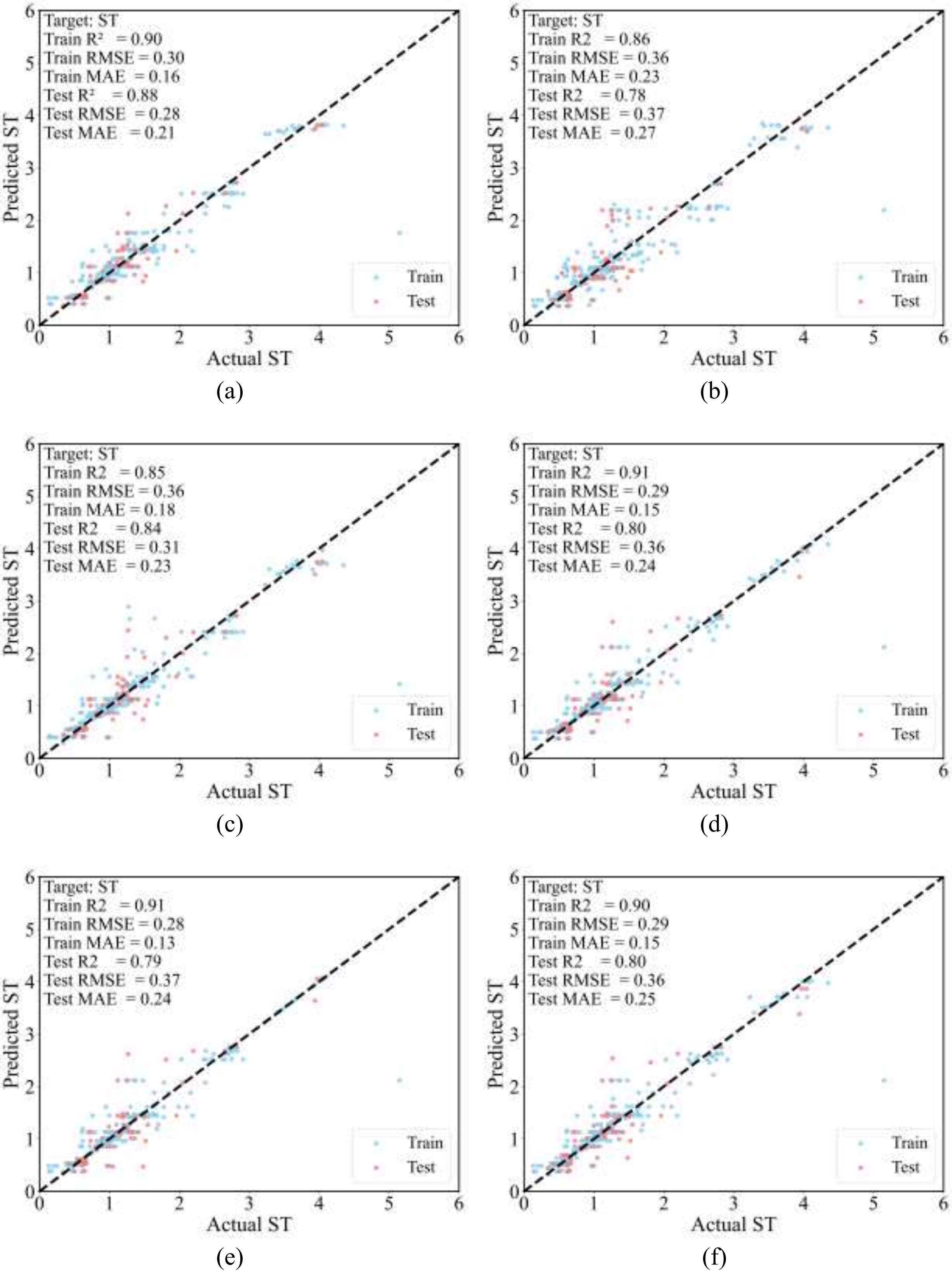
Target: ST
Train R² = 0.90
Train RMSE = 0.30
Train MAE = 0.16
Test R² = 0.88
Test RMSE = 0.28
Test MAE = 0.21
Predicted ST
Actual ST
Train
Test
(a)
Target: ST
Train R2 = 0.86
Train RMSE = 0.36
Train MAE = 0.23
Test R2 = 0.78
Test RMSE = 0.37
Test MAE = 0.27
Predicted ST
Actual ST
Train
Test
(b)
Target: ST
Train R2 = 0.85
Train RMSE = 0.36
Train MAE = 0.18
Test R2 = 0.84
Test RMSE = 0.31
Test MAE = 0.23
Predicted ST
Actual ST
Train
Test
(c)
Target: ST
Train R2 = 0.91
Train RMSE = 0.29
Train MAE = 0.15
Test R2 = 0.80
Test RMSE = 0.36
Test MAE = 0.24
Predicted ST
Actual ST
Train
Test
(d)
Target: ST
Train R2 = 0.91
Train RMSE = 0.28
Train MAE = 0.13
Test R2 = 0.79
Test RMSE = 0.37
Test MAE = 0.24
Predicted ST
Actual ST
Train
Test
(e)
Target: ST
Train R2 = 0.90
Train RMSE = 0.29
Train MAE = 0.15
Test R2 = 0.80
Test RMSE = 0.36
Test MAE = 0.25
Predicted ST
Actual ST
Train
Test
(f)

Fig. 7. Performance of ML models for ST prediction on the training and testing sets: (a) TabPFN; (b) ANN; (c) SVR; (d) RF (e) XGBoost; (f) LightGBM.

The multi-metric predictive performance of the six models for ST is summarized in **Fig. 8**. As shown in **Fig. 8(a)**, TabPFN achieved the best overall performance across the five evaluation metrics, showing clear advantages in RMSE, MAPE, MAE, MAD, and $R^2$. SVR ranked second and also exhibited relatively balanced predictive capability. XGBoost, RF, and LightGBM showed moderate overall performance, while ANN obtained the lowest scores among the six models. This trend is further confirmed by the composite scores in **Fig. 8(b)**, where TabPFN reached the highest score of 0.91, followed by SVR (0.66). Overall, the results indicate that TabPFN is the most effective model for ST prediction in the present dataset, whereas the other models still provide acceptable but comparatively weaker performance.

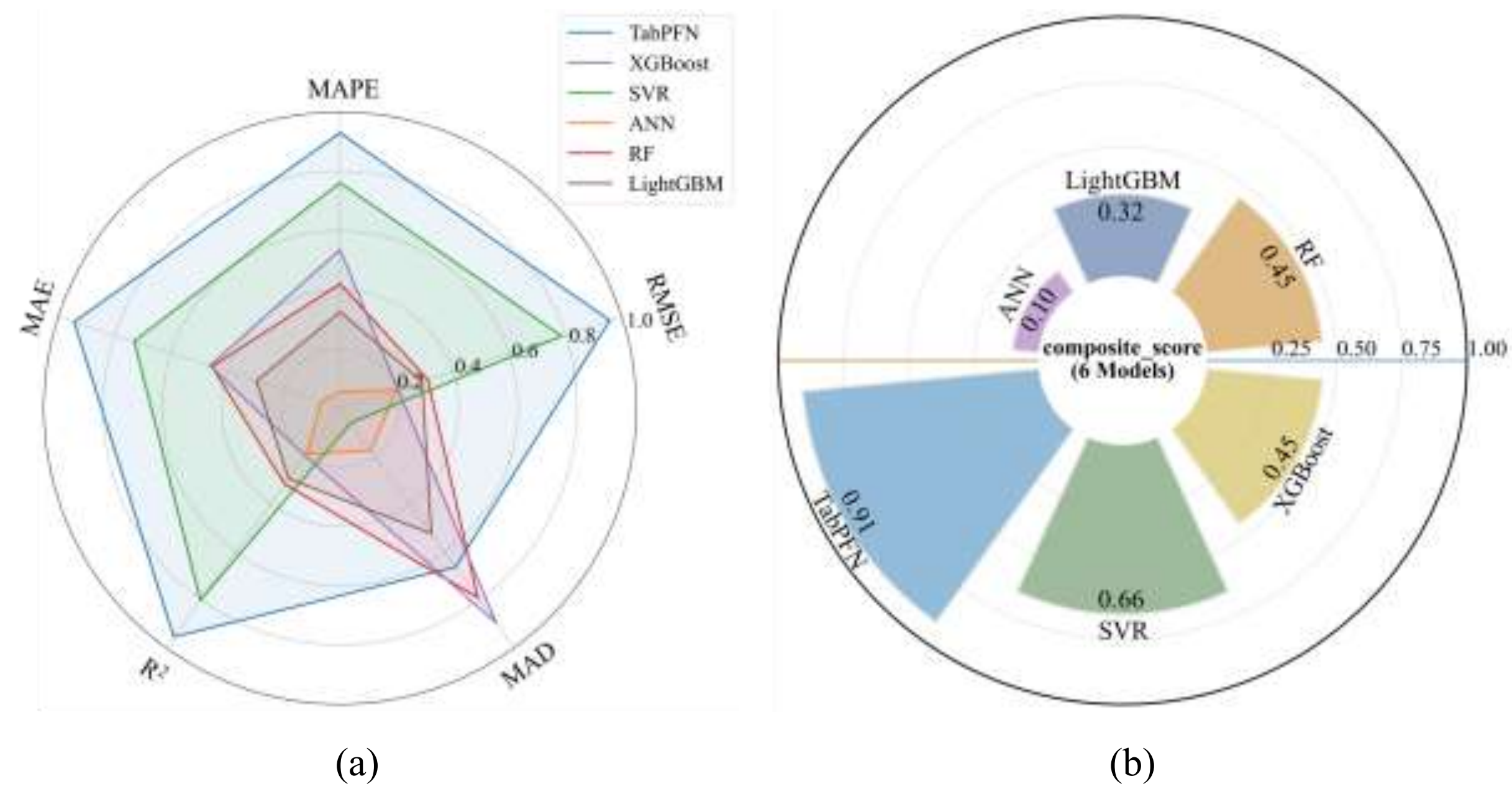

(a) (b)

Fig. 8. Performance evaluation of different models for ST prediction: (a) radar chart based on five evaluation metrics; (b) composite score integrating the five metrics.

### 3.3. Local interpretability based on SHAP

To provide a case-level interpretation of the prediction behavior of the six machine-learning models, a randomly chosen sample was examined in detail. The corresponding feature values were AC = 4.6 wt.%, Pe = 72 (0.1 mm), Du = 137 cm, SP = 51 °C, AV = 4.48%, VMA = 17.51%, VFA = 71.97%, Ag2.36 = 56%, Ag4.75 = 69%, Ag9.5 = 86%, FT = No_fiber, FC = 0 wt.%, FL = 0 mm, TS = 0 MPa, and ST = 0.93 MPa. As shown in the SHAP force plots in **Fig. 9**, all six models predict the selected sample at values lower than their corresponding base values, indicating an overall downward shift from the average prediction. Although the dominant contributing features vary across models, some consistent patterns can still be observed. In particular, FT (No_fiber) shows a negative contribution in all six models, while VMA, Ag4.75, Ag2.36, and Pe also tend to reduce the predicted ST in most cases. By contrast, the positive contributions are more model-dependent, with features such as Du, AC, SP, AV, FC, TS, and FL increasing the predicted ST in some models. The final ST values predicted by TabPFN, ANN, SVR, RF, XGBoost, and LightGBM were 0.94, 1.04, 1.01, 0.96, 0.97, and 0.97 MPa, respectively. Given that the actual ST value of the selected sample is 0.93 MPa, the corresponding relative errors are 1.08%, 11.83%, 8.60%, 3.23%, 4.30%, and 4.30%, respectively. Among the six models, TabPFN yields the closest prediction to the actual value, while RF, XGBoost, and LightGBM also maintain prediction errors within 5%, indicating satisfactory local predictive accuracy for this sample.

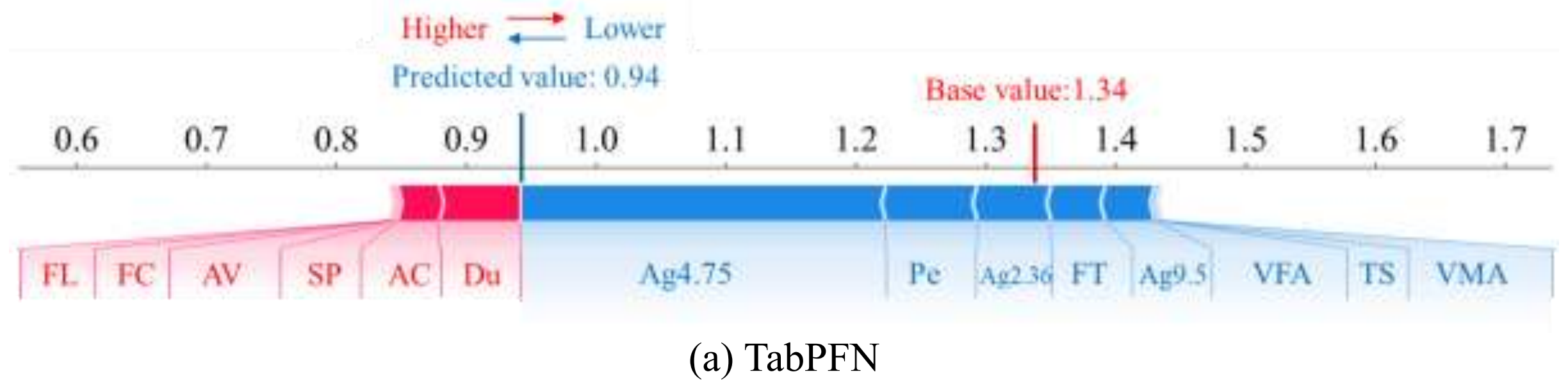

(a) TabPFN

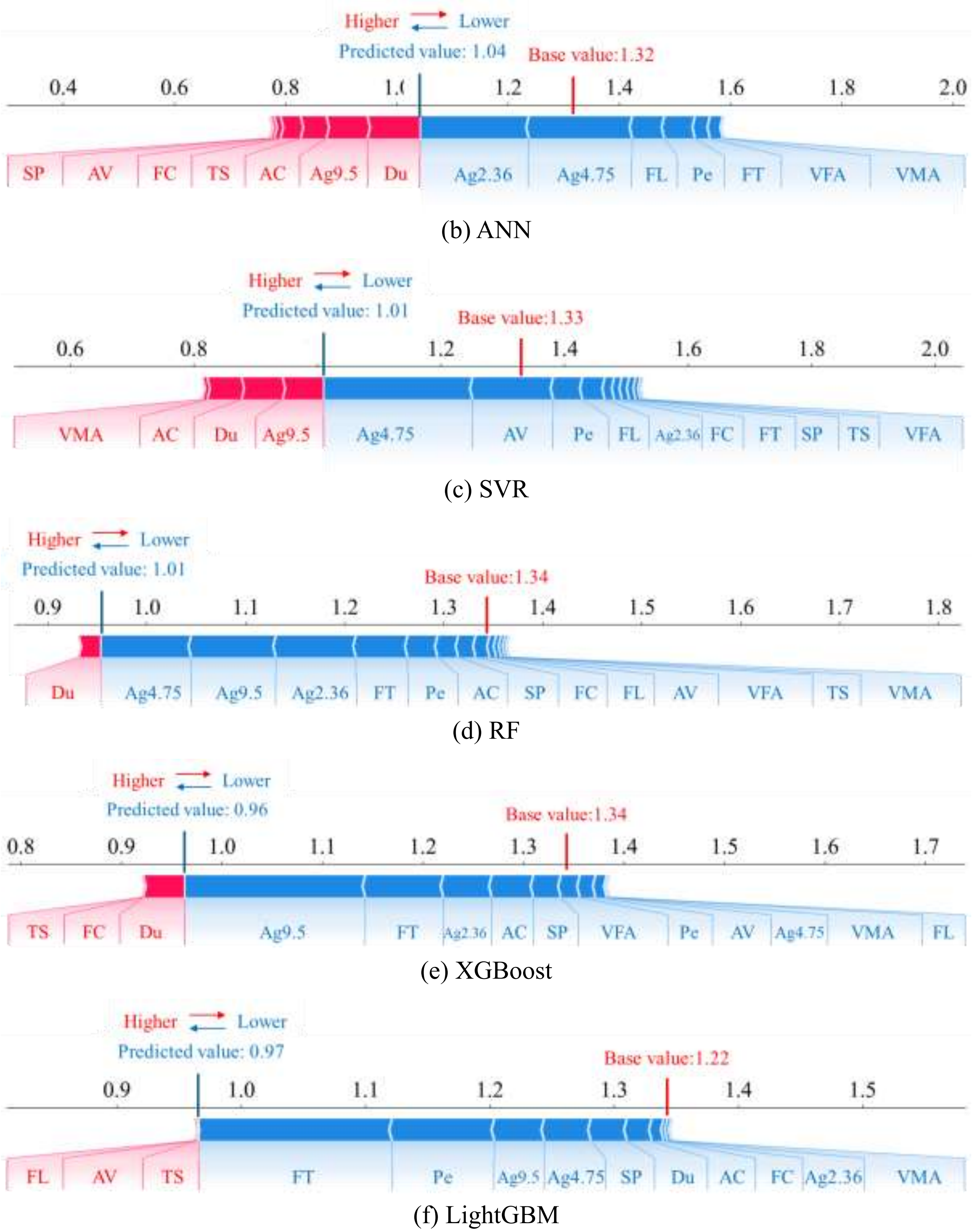

(b) ANN

(c) SVR

(d) RF

(e) XGBoost

(f) LightGBM

Fig. 9. Illustration of the prediction behavior of the six machine-learning models. Red bars represent positive effects, whereas blue bars represent negative effects.

## 3.4. Global interpretability based on SHAP

### *3.4.1. Contribution of individual features*

In addition to local explanation for a single sample, SHAP can also be used to quantify the overall influence of input variables on ST prediction across the full dataset. **Fig. 10** presents the global SHAP interpretation results of the TabPFN model, selected because of its superior predictive accuracy. The pie chart reports the proportional contribution of each feature based on the sum of absolute SHAP values, while the beeswarm plot further illustrates the distribution and polarity of feature effects for all samples. In the beeswarm plot, each point represents a sample, and the color scale indicates the feature value from low to high. The horizontal axis corresponds to the SHAP value, where positive and negative values denote increasing and decreasing effects on the predicted ST, respectively. Features are ordered by mean absolute SHAP value, allowing direct comparison of their overall importance. The results show that Ag4.75 is the most influential predictor, with AC, Ag9.5, FT, Pe, and Du also contributing substantially, whereas VMA, FC, FL, and TS play relatively minor roles in the model output. The beeswarm plots of the remaining five models are shown in **Fig. C1** of **Appendix C**.

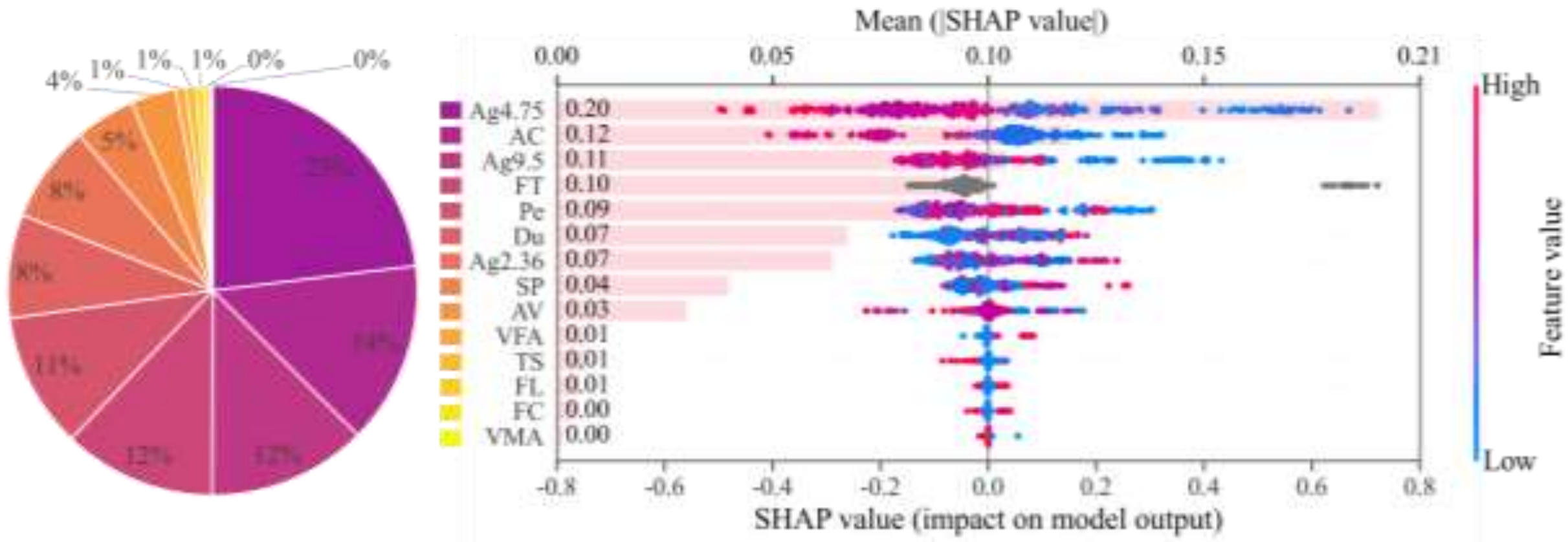

Fig. 10. SHAP analysis of the TabPFN predictions.

As reflected by the pie charts in **Fig. 10** and **Fig. C1**, the relative importance of individual variables is not exactly identical across models; however, a clear overall pattern can still be observed. Based on the averaged percentages summarized in **Table 4**, Ag9.5, FT, Ag4.75, AC, and Du are identified as high-impact variables, each contributing more than 10% on average. Pe, Ag2.36, SP, and AV fall into the medium-impact group, with average contributions between 5% and 10%. In contrast, FL, FC, TS, VFA, and VMA exhibit average contribution percentages below 5%, indicating relatively limited influence on the model output. In aggregate, high-, medium-, and low-impact features account for 65.5%, 26.5%, and 7.9% of the total influence, respectively, highlighting the dominant role of a small subset of variables in determining ST prediction.

Table 4. Ranking of variable importance for ST prediction

| Variables | Proportion of models (%) | | | | | | Mean (%) |
|---|---|---|---|---|---|---|---|
| | TabPFN | ANN | SVR | RF | XGBoost | LightGBM | |
| Ag9.5 | 12 | 11 | 15 | 22 | 33 | 20 | 18.8 |
| FT | 12 | 7 | 7 | 12 | 14 | 25 | 12.8 |
| Ag4.75 | 23 | 15 | 13 | 11 | 1 | 7 | 11.7 |
| AC | 14 | 15 | 11 | 10 | 8 | 9 | 11.2 |
| Du | 8 | 10 | 9 | 10 | 17 | 12 | 11 |
| Pe | 11 | 7 | 9 | 8 | 7 | 11 | 8.8 |
| Ag2.36 | 8 | 6 | 7 | 8 | 3 | 4 | 6 |
| SP | 5 | 8 | 7 | 6 | 7 | 3 | 6 |
| AV | 4 | 5 | 7 | 5 | 6 | 7 | 5.7 |
| FL | 1 | 8 | 7 | 2 | 0 | 0 | 3 |
| FC | 0 | 1 | 4 | 4 | 1 | 1 | 1.8 |
| TS | 1 | 4 | 2 | 1 | 0 | 0 | 1.3 |
| VFA | 1 | 2 | 1 | 1 | 3 | 0 | 1.3 |
| VMA | 0 | 1 | 1 | 0 | 0 | 1 | 0.5 |

### *3.4.2. Feature-wise dependence analysis*

To examine the variation patterns and possible threshold behaviors of key variables affecting ST, **Fig. 11** displays the SHAP dependence plots of all input features obtained from the TabPFN model, thereby revealing how each parameter influences the ST of asphalt concrete.

Based on **Table 4**, the nine variables shown in **Fig. 11** (Ag9.5, FT, Ag4.75, AC, Du, Pe, Ag2.36, SP, and AV) were selected for dependence analysis because they comprise all high- and medium-impact features and jointly explain 92.0% of the total average SHAP contribution for ST prediction. Their SHAP dependence plots were fitted with LOWESS curves and accompanied by ±0.5 standard deviation error bands to characterize nonlinear trends and local uncertainty. This visualization facilitates parametric interpretation of the relationships between feature values and their corresponding SHAP effects on the predicted splitting strength. In contrast, the five low-impact variables (FL, FC, TS, VFA, and VMA), whose combined contribution is only 7.9%, were excluded from further discussion. Since FT is a categorical descriptor, its plot is presented as grouped scatter distributions for different fiber types rather than as a continuous fitted curve.

When the baseline ST value is 1.34 (see **Fig. 9(a)**), the SHAP dependence plots in **Fig. 11** show that the nine dominant variables can be classified into three types according to their influence patterns. First, Ag9.5, Ag4.75, AC, and AV exhibit overall negative correlations with ST, as shown in **Fig. 11(a), (c), (d),** and **(i)**, suggesting that larger values of these variables are generally unfavorable for improving splitting strength. From a physical perspective, this pattern can be attributed to the progressive weakening of the internal load-carrying structure of the mixture. Higher Ag9.5 and Ag4.75 passing rates generally indicate a gradation shift toward a less effective aggregate skeleton and weaker interlocking action [80]. Likewise, excessive AC may lead to an overly binder-rich system, in which thick asphalt films reduce the contribution of aggregate interlock to tensile resistance [81]. For AV, its negative effect is more direct, since a higher void content increases internal discontinuities and stress concentration, thus making crack initiation and propagation easier during the splitting process [82]. Second, Du shows an overall positive correlation with ST, as illustrated in **Fig. 11(e)**, with its SHAP contribution becoming positive after a certain threshold. This trend is physically reasonable because higher ductility indicates a greater ability of the asphalt binder to accommodate tensile deformation and dissipate fracture energy, thereby reducing stress concentration and delaying crack propagation during the splitting process [83]. Third, Pe, Ag2.36, and SP display non-monotonic effects on ST, as shown in **Fig. 11(f), (g),** and **(h)**, indicating that their contributions vary across different value ranges. This non-monotonic behavior is physically plausible because Pe, Ag2.36, and SP affect ST through structural and binder-property balance rather than through a simple linear mechanism, so only certain value ranges are favorable for resisting splitting failure [84]-[86]. For the categorical variable FT in **Fig. 11(b)**, the SHAP distribution reveals a pronounced category-dependent pattern rather than a uniform fiber-reinforcement effect. Polyester fiber is associated with the highest positive contribution, whereas the other fiber categories show lower or near-neutral SHAP values. Given the imbalance among fiber categories in the present database, these results should be interpreted as dataset-dependent relative model contributions rather than definitive judgments on the intrinsic effectiveness of individual fiber types. According to these relationships, the following thresholds or favorable ranges are suggested for enhancing ST performance: Ag9.5 < 66.8%, Ag4.75 < 45.0%, AC < 5.4 wt.%, AV < 3.6%, Du > 134.7 cm, Pe < 60 or Pe > 86.7 (0.1 mm), 37.0% < Ag2.36 < 51.5%, and SP < 45.6 °C or SP > 55.6 °C.

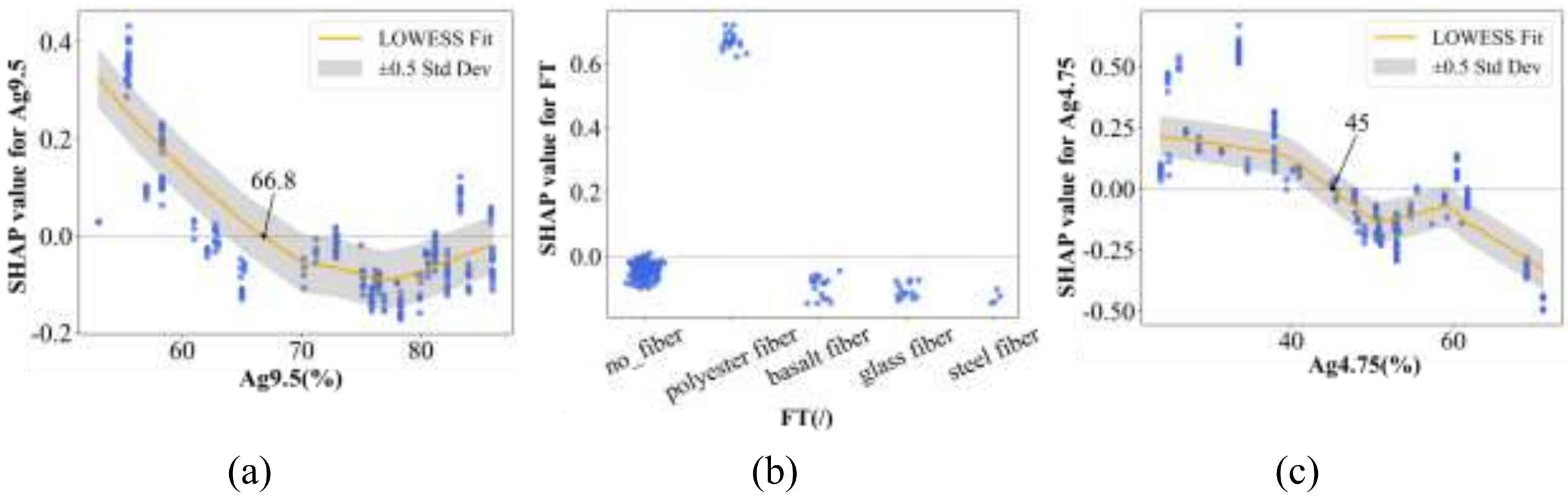

(a) (b) (c)

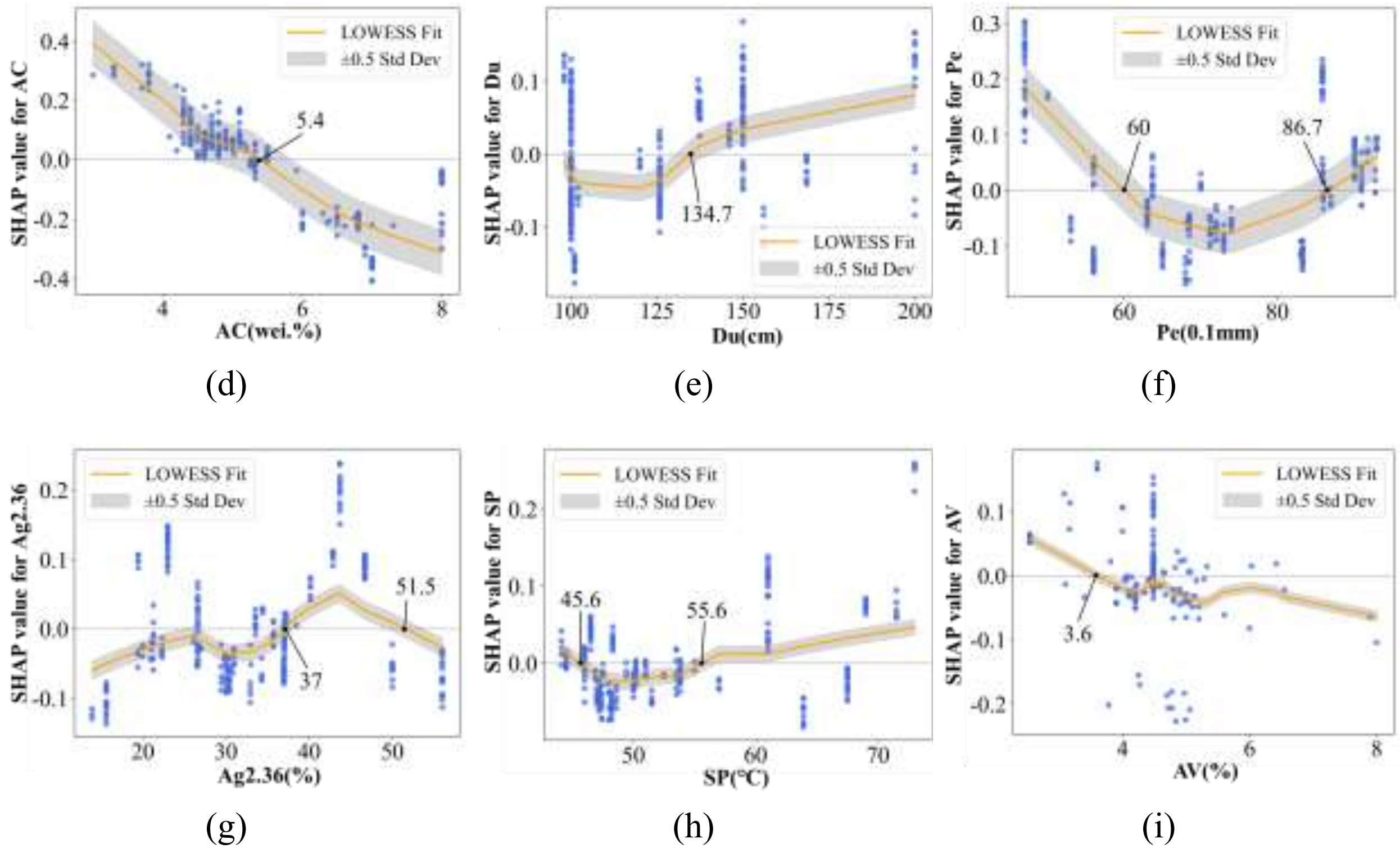

Fig. 11. SHAP dependence plots of the nine key input variables for ST prediction based on the TabPFN model: (a) Ag9.5; (b) FT; (c) Ag4.75; (d) AC; (e) Du; (f) Pe; (g) Ag2.36; (h) SP; (i) AV.

## 4. Graphical user interface platform

As shown in **Fig. 12**, the graphical user interface (GUI) was developed in Python based on the Streamlit framework. It enables users to enter 14 variables associated with asphalt mixture composition, aggregate gradation, and fiber characteristics for splitting strength (ST) prediction. After entering the required parameters, the current raw input is displayed in tabular form, and the user can click the prediction button to obtain the predicted ST value generated by the deployed pre-trained model. In addition to prediction, the platform provides SHAP-based interpretability analysis for the current sample. As illustrated in **Fig. 12**, the GUI presents a waterfall plot to visualize how individual input features contribute positively or negatively to the final ST prediction, thereby improving the transparency and interpretability of the model. Therefore, the developed platform serves not only as a practical prediction tool, but also as an interpretable decision-support interface for asphalt concrete splitting strength evaluation. The platform is available at https://st-gui-app-nlj7snzfjkvdaqfqf4yvkv.streamlit.app/

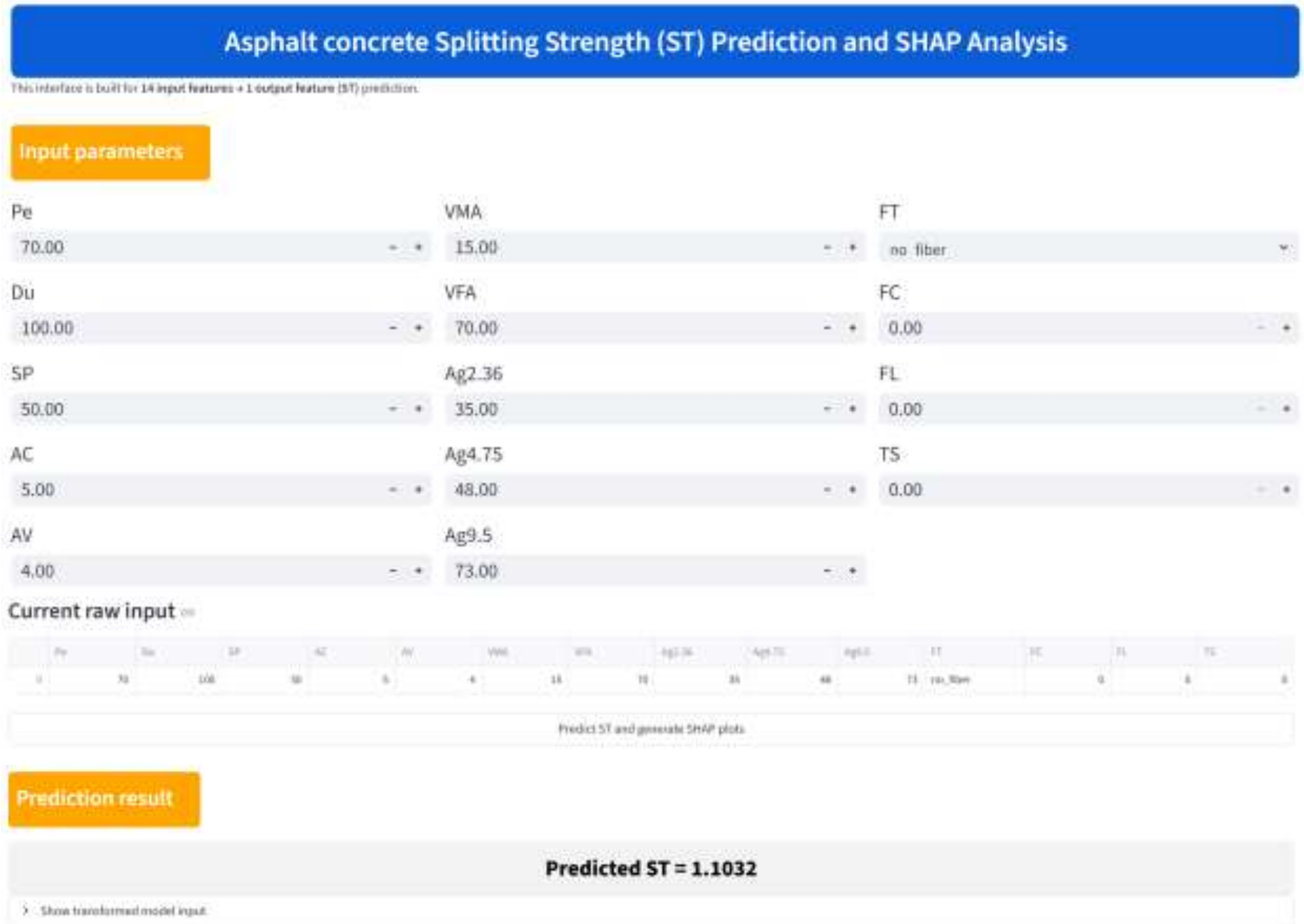

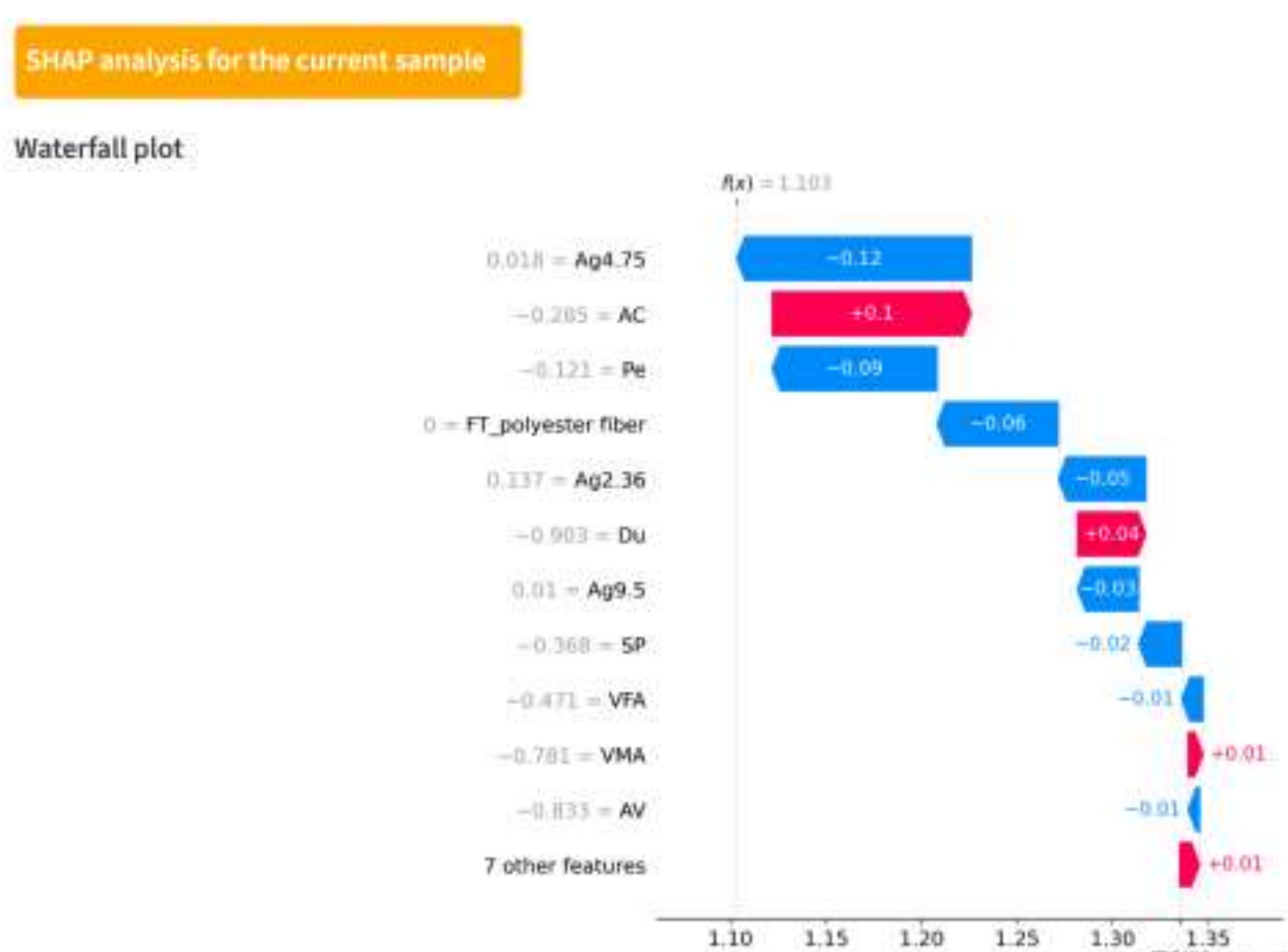

Fig. 12. GUI for asphalt concrete splitting strength (ST) prediction and SHAP-based interpretation.

## 5. Limitations and future work

### 5.1. Overall effectiveness

This study demonstrates the overall effectiveness of an interpretable data-driven framework for ST prediction of asphalt concrete by integrating dataset preprocessing, multi-model learning, hyperparameter optimization, SHAP-based explanation, and a user-oriented GUI platform. The results presented in Section 3 indicate that all six machine learning models achieved acceptable prediction accuracy. Among them, TabPFN showed the strongest overall performance on the testing set and obtained the highest composite score, demonstrating that the proposed framework can effectively learn the complex nonlinear relationships between asphalt-, aggregate-, and fiber-related variables and the resulting splitting strength. Beyond predictive accuracy, the SHAP analysis further improves the engineering usefulness of the framework by identifying the dominant variables and clarifying their different influence patterns, including negative, positive, non-monotonic, and category-dependent effects. Therefore, the main value of the present work lies not only in accurate ST prediction, but also in providing interpretable parameter-level guidance for mixture design and offering a practical basis for future digital design tools for asphalt concrete.

### 5.2. Challenges and limitations

Despite the promising results, several limitations remain. Since the database was assembled from different published studies, variations in material sources, mix design details, specimen preparation, testing conditions, and reporting practices are unavoidable, which may affect the consistency and transferability of the developed models. In addition, the distribution of fiber types is not fully balanced, and this may reduce the reliability of the category-specific patterns identified for FT. It should therefore be emphasized that the SHAP results for FT reflect relative model contributions under the present dataset rather than definitive judgments on the intrinsic effectiveness of each fiber type. Moreover, although SHAP provides useful interpretability, it describes model-learned associations rather than causal mechanisms, meaning that the reported thresholds and favorable intervals should be treated as empirical guidance. Finally, the present framework is limited to ST prediction and has not yet been extended to long-term service behavior, environmental coupling effects, or full engineering life-cycle evaluation.

### 5.3. Opportunities for future research

Future studies should further strengthen the data basis and expand the application range of the current framework. An important direction for future work is to establish a larger and more standardized asphalt concrete database with more balanced fiber categories, more complete reporting of material properties, and more consistent testing protocols, so that model training and interpretation can become more robust and transferable. On this basis, future studies can extend the framework from single-property prediction to multi-objective design by jointly considering strength, cracking resistance, rutting resistance, durability, and sustainability indicators, thereby supporting more comprehensive optimization of asphalt mixtures. In parallel, combining SHAP with stronger causal analysis, uncertainty quantification, and external validation from laboratory or field data would make the design recommendations more credible and engineering-oriented. Finally, the current GUI can be further upgraded into a continuously updated intelligent platform in which new data are periodically incorporated, models are retrained, and users can obtain not only ST predictions but also dynamic recommendation ranges for mixture parameters under different engineering scenarios.

## 6. Conclusion

This study established an interpretable machine-learning framework to predict the splitting strength (ST) of asphalt concrete using a literature-based database with fourteen input variables associated with asphalt properties, aggregate gradation, and fiber characteristics. Six machine learning models were established and compared, and the framework further integrated hyperparameter optimization, SHAP-based interpretation, and a GUI platform to improve both predictive capability and engineering usability. Based on the findings of this study, the following conclusions can be drawn.

- All six machine learning models demonstrated satisfactory capability for ST prediction, confirming that ML is effective in capturing the nonlinear relationships between mixture design variables and splitting strength. Among them, TabPFN achieved the best overall predictive performance on the testing set, with the lowest RMSE of 0.28, the highest $R^2$ of 0.88, and the highest composite score of 0.91. SVR ranked second overall, while XGBoost, RF, and LightGBM showed moderate but still acceptable predictive performance. These results indicate that TabPFN is the most suitable model for ST prediction in the present dataset.

- The SHAP analysis showed that the prediction of ST is mainly governed by a limited number of dominant variables. Based on the average feature contributions across the six models, Ag9.5, FT, Ag4.75, AC, and Du were identified as high-impact variables, while Pe, Ag2.36, SP, and AV were classified as medium-impact variables. Together, these nine variables accounted for 92.0% of the total average SHAP contribution, whereas FL, FC, TS, VFA, and VMA had relatively minor influence. In addition, the SHAP force-plot analysis for a representative sample showed that TabPFN provided the closest prediction to the actual ST value, further confirming its strong local interpretability and predictive reliability.
- The SHAP dependence analysis further revealed that the dominant variables exhibit different influence patterns on ST, including overall negative correlations, positive correlations, non-monotonic effects, and category-dependent effects. Specifically, Ag9.5, Ag4.75, AC, and AV showed overall negative correlations with ST; Du showed an overall positive correlation; and Pe, Ag2.36, and SP exhibited non-monotonic relationships. For the categorical feature FT, polyester fiber showed a comparatively stronger positive contribution to ST than the other fiber types in the present dataset under the current data conditions. Based on the dependence analysis, the favorable ranges for improving ST were identified as Ag9.5 < 66.8%, Ag4.75 < 45.0%, AC < 5.4 wt.%, AV < 3.6%, Du > 134.7 cm, Pe < 60 or > 86.7 (0.1 mm), 37.0% < Ag2.36 < 51.5%, and SP < 45.6 °C or > 55.6 °C.
- Beyond model construction and interpretation, this study also established a GUI platform to enhance the accessibility and applicability of the developed framework. By integrating prediction and SHAP-based explanation into a user-oriented interface, the platform provides a practical tool for estimating ST and understanding the role of individual design variables. Overall, the proposed framework offers not only accurate prediction of splitting strength, but also interpretable guidance for mixture design, thereby demonstrating the potential of explainable artificial intelligence in the intelligent design and optimization of asphalt concrete.

## Data Availability

Data will be made available on request.

## Declaration of Competing Interest

The authors declare no competing financial interests or personal affiliations that could have influenced the results presented in this manuscript.

## CRediT Authorship Contribution Statement

**Xxxxxxx**

## Acknowledgement

xxxxx

## Ethical Approval

The results/data/figures in this manuscript have not been published elsewhere, nor are they under consideration by another publisher.

## Appendix A. Data description

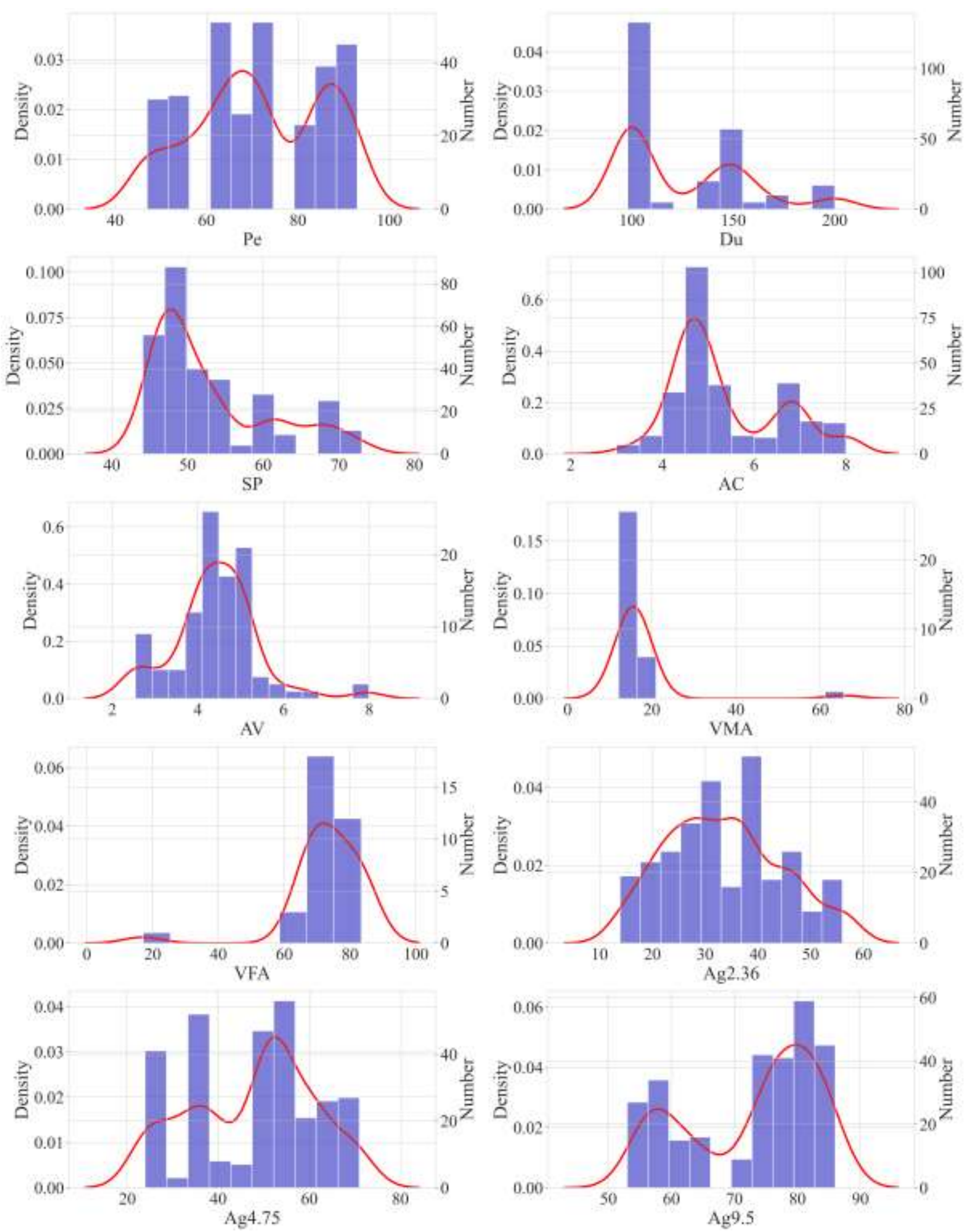

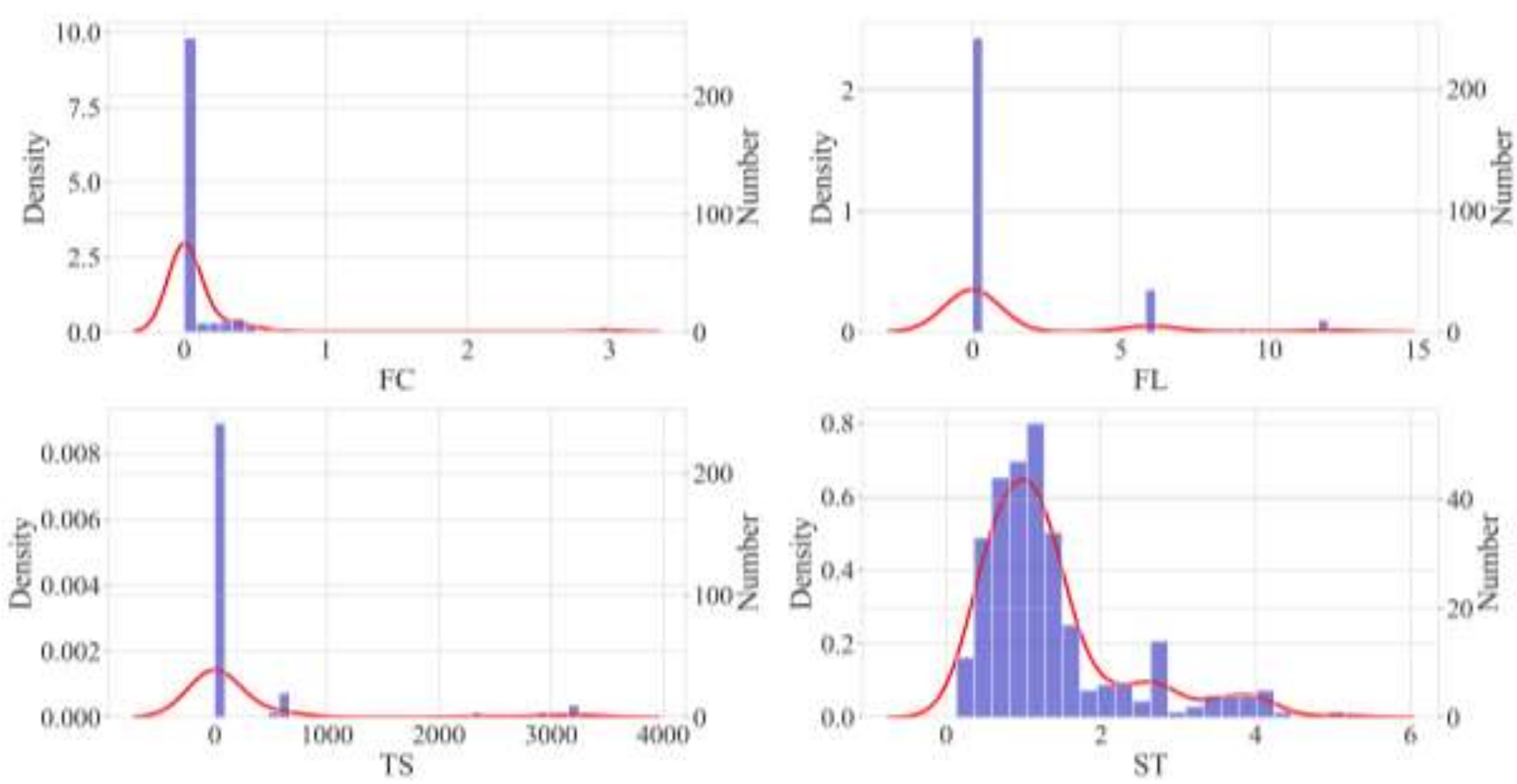

Fig. A1. Distribution of fourteen input features and one output feature of the dataset.

.

## Appendix B. Model configuration and performance evaluation

Table B1. Hyperparameter combinations adopted for each model

| Models | Hyperparameters |
|---|---|
| TabPFN | default hyperparameters |
| ANN | n= 2<br>hidden_layer_sizes = (128, 64)<br>learning_rate_init = 0.01<br>batch_size = 151<br>activation = relu<br>solver = adam<br>validation_fraction = 0.1<br>early_stopping=True |
| SVR | C = 5.48<br>gamma = 0.14<br>epsilon = 0.24<br>kernel = rbf |
| RF | n_estimators = 370<br>max_depth = 13<br>min_samples_split = 2<br>min_samples_leaf = 1<br>max_features = log2<br>bootstrap = True |
| XGBoost | n_estimators = 207<br>learning_rate = 0.09<br>max_depth = 6<br>objective = reg:squarederror<br>tree_method = hist |
| LightGBM | n_estimators = 477<br>learning_rate = 0.05<br>max_depth = 7<br>min_child_samples= 12<br>reg_alpha = 0.07<br>reg_lambda = 0.04<br>num_leaves = 57 |

Table B2. Performance metrics of the models for ST prediction on the testing set

| Model | Metrics | | | | |
|---|---|---|---|---|---|
| | RMSE | MAE | MAPE | MAD | $R^2$ |
| TabPFN | 0.28 | 0.21 | 18.01 | 0.14 | 0.88 |
| ANN | 0.37 | 0.27 | 24.87 | 0.16 | 0.78 |
| SVR | 0.31 | 0.23 | 19.81 | 0.17 | 0.84 |
| RF | 0.36 | 0.24 | 21.69 | 0.14 | 0.80 |
| XGBoost | 0.37 | 0.24 | 21.11 | 0.13 | 0.79 |
| LightGBM | 0.36 | 0.25 | 22.21 | 0.14 | 0.80 |

## Appendix C. SHAP analysis demonstration

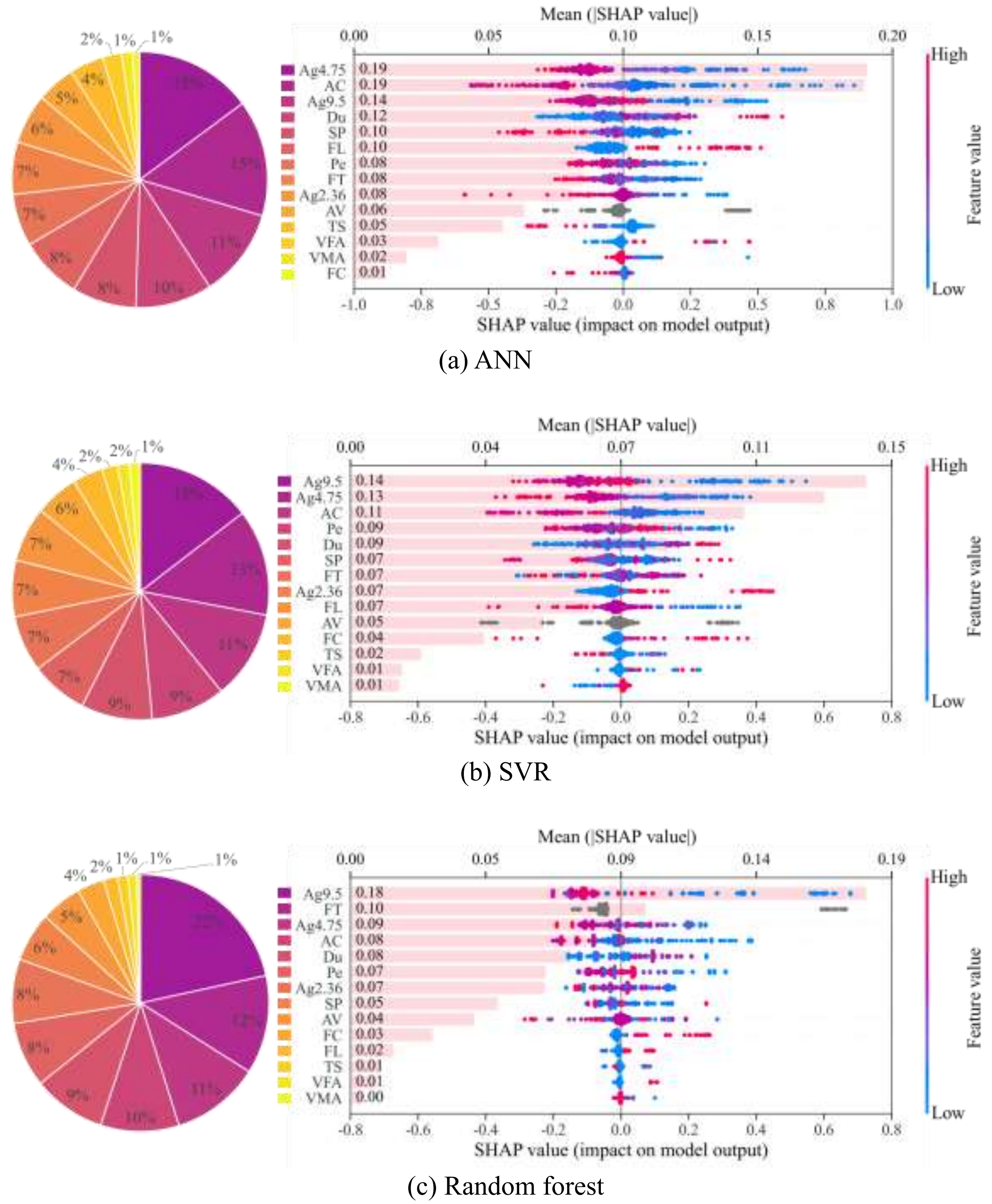

(a) ANN

(b) SVR

(c) Random forest

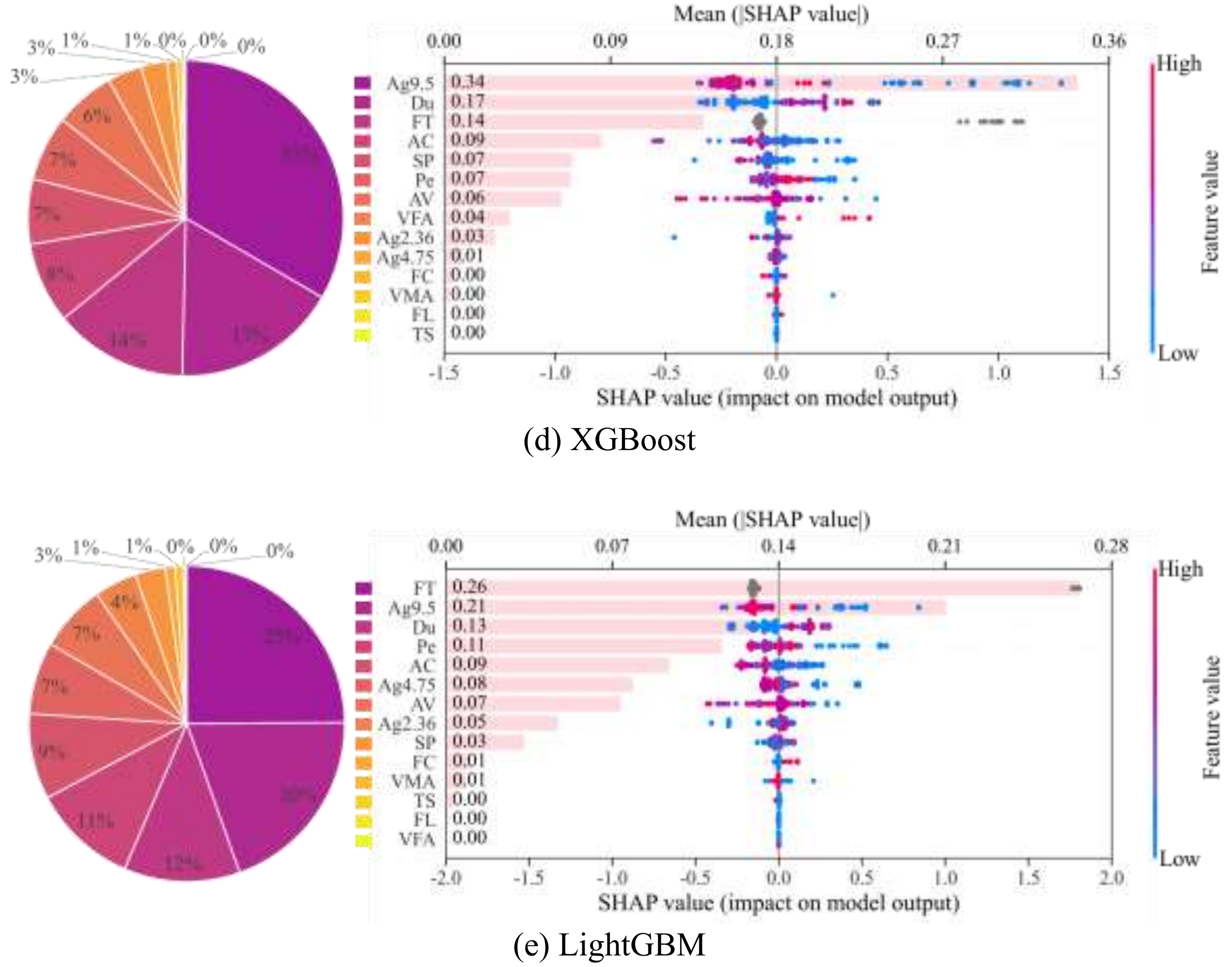

Fig. C1. Summary plot of SHAP-based interpretation of the outputs from the remaining five machine learning models: (a) ANN; (b) SVR; (c) RF; (d) XGBoost; (e) LightGBM